%% file: neurips_2024.tex
\documentclass{article}

\PassOptionsToPackage{numbers, compress}{natbib}

\usepackage[preprint]{neurips_2024}

\usepackage[utf8]{inputenc} 
\usepackage[T1]{fontenc}    
\usepackage{hyperref}       
\usepackage{url}            
\usepackage{booktabs}       
\usepackage{amsfonts}       
\usepackage{nicefrac}       
\usepackage{microtype}      
\usepackage{xcolor}         
\usepackage{amsmath}
\usepackage{amssymb}
\usepackage{graphicx}
\usepackage{wrapfig}
\usepackage{subcaption}
\usepackage{hyperref}
\usepackage{float}
\usepackage{makecell}
\usepackage{xspace}
\usepackage{enumitem}

\newcommand{\method}{\texttt{DILA}\xspace}
\usepackage[ruled,vlined, linesnumbered]{algorithm2e}
\title{\method: Dictionary Label Attention for Mechanistic Interpretability in High-dimensional Multi-label Medical Coding Prediction}

\author{%
  John Wu \\
  University of Illinois Urbana-Champaign \\
  \texttt{johnwu3@illinois.edu} \\
  \And
  David Wu \\
  Vanderbilt University \\
  \texttt{David.h.wu@vanderbilt.edu} \\
  \And
  Jimeng Sun \\
  University of Illinois Urbana-Champaign \\
  \texttt{jimeng@illinois.edu} \\
}

\begin{document}

\maketitle

\begin{abstract}
 Predicting high-dimensional or extreme multilabels, such as in medical coding, requires both accuracy and interpretability. Existing works often rely on local interpretability methods, failing to provide comprehensive explanations of the overall mechanism behind each label prediction within a multilabel set. We propose a mechanistic interpretability module called DIctionary Label Attention (\method) that disentangles uninterpretable dense embeddings into a sparse embedding space, where each nonzero element (a dictionary feature) represents a globally learned medical concept. Through human evaluations, we show that our sparse embeddings are more human understandable than its dense counterparts by at least 50 percent. Our automated dictionary feature identification pipeline, leveraging large language models (LLMs), uncovers thousands of learned medical concepts by examining and summarizing the highest activating tokens for each dictionary feature. We represent the relationships between dictionary features and medical codes through a sparse interpretable matrix, enhancing the mechanistic and global understanding of the model's predictions while maintaining competitive performance and scalability without extensive human annotation.
     
\end{abstract}
\footnote{Code available at: https://github.com/jhnwu3/DILA}
\input{sections/1introduction}

\input{sections/2related_work}
\input{sections/3method}
\input{sections/4results_discussion}
\input{sections/5conclusion}

\newpage
\bibliographystyle{acm}
\bibliography{ref}





\appendix

\section{Appendix / supplemental material}
\input{sections/6appendix}


\newpage
\section*{NeurIPS Paper Checklist}

\begin{enumerate}

\item {\bf Claims}
    \item[] Question: Do the main claims made in the abstract and introduction accurately reflect the paper's contributions and scope?
    \item[] Answer:  \answerYes{}
    \item[] Justification:  Our paper shares human evaluation results and other results to showcase the scalability of our mechanistic interpretability method. 
    \item[] Guidelines:
    \begin{itemize}
        \item The answer NA means that the abstract and introduction do not include the claims made in the paper.
        \item The abstract and/or introduction should clearly state the claims made, including the contributions made in the paper and important assumptions and limitations. A No or NA answer to this question will not be perceived well by the reviewers. 
        \item The claims made should match theoretical and experimental results, and reflect how much the results can be expected to generalize to other settings. 
        \item It is fine to include aspirational goals as motivation as long as it is clear that these goals are not attained by the paper. 
    \end{itemize}

\item {\bf Limitations}
    \item[] Question: Does the paper discuss the limitations of the work performed by the authors?
    \item[] Answer: \answerYes{} 
    \item[] Justification: We discuss many of the limitations of our method in section \ref{sec: Discussion}. We also go into further details about the challenges of using conventional faithfulness metrics in our appendix. 
    \item[] Guidelines:
    \begin{itemize}
        \item The answer NA means that the paper has no limitation while the answer No means that the paper has limitations, but those are not discussed in the paper. 
        \item The authors are encouraged to create a separate "Limitations" section in their paper.
        \item The paper should point out any strong assumptions and how robust the results are to violations of these assumptions (e.g., independence assumptions, noiseless settings, model well-specification, asymptotic approximations only holding locally). The authors should reflect on how these assumptions might be violated in practice and what the implications would be.
        \item The authors should reflect on the scope of the claims made, e.g., if the approach was only tested on a few datasets or with a few runs. In general, empirical results often depend on implicit assumptions, which should be articulated.
        \item The authors should reflect on the factors that influence the performance of the approach. For example, a facial recognition algorithm may perform poorly when image resolution is low or images are taken in low lighting. Or a speech-to-text system might not be used reliably to provide closed captions for online lectures because it fails to handle technical jargon.
        \item The authors should discuss the computational efficiency of the proposed algorithms and how they scale with dataset size.
        \item If applicable, the authors should discuss possible limitations of their approach to address problems of privacy and fairness.
        \item While the authors might fear that complete honesty about limitations might be used by reviewers as grounds for rejection, a worse outcome might be that reviewers discover limitations that aren't acknowledged in the paper. The authors should use their best judgment and recognize that individual actions in favor of transparency play an important role in developing norms that preserve the integrity of the community. Reviewers will be specifically instructed to not penalize honesty concerning limitations.
    \end{itemize}

\item {\bf Theory Assumptions and Proofs}
    \item[] Question: For each theoretical result, does the paper provide the full set of assumptions and a complete (and correct) proof?
    \item[] Answer: \answerNA{} 
    \item[] Justification: There are no theoretical results here. 
    \item[] Guidelines:
    \begin{itemize}
        \item The answer NA means that the paper does not include theoretical results. 
        \item All the theorems, formulas, and proofs in the paper should be numbered and cross-referenced.
        \item All assumptions should be clearly stated or referenced in the statement of any theorems.
        \item The proofs can either appear in the main paper or the supplemental material, but if they appear in the supplemental material, the authors are encouraged to provide a short proof sketch to provide intuition. 
        \item Inversely, any informal proof provided in the core of the paper should be complemented by formal proofs provided in appendix or supplemental material.
        \item Theorems and Lemmas that the proof relies upon should be properly referenced. 
    \end{itemize}

    \item {\bf Experimental Result Reproducibility}
    \item[] Question: Does the paper fully disclose all the information needed to reproduce the main experimental results of the paper to the extent that it affects the main claims and/or conclusions of the paper (regardless of whether the code and data are provided or not)?
    \item[] Answer: \answerYes{} 
    \item[] Justification: We do our best to record much of the hyperparameter training details in our Appendix. We note that physionet prevents us from directly sharing the dataset, but it is open source once users apply and do some privacy training. 
    \item[] Guidelines:
    \begin{itemize}
        \item The answer NA means that the paper does not include experiments.
        \item If the paper includes experiments, a No answer to this question will not be perceived well by the reviewers: Making the paper reproducible is important, regardless of whether the code and data are provided or not.
        \item If the contribution is a dataset and/or model, the authors should describe the steps taken to make their results reproducible or verifiable. 
        \item Depending on the contribution, reproducibility can be accomplished in various ways. For example, if the contribution is a novel architecture, describing the architecture fully might suffice, or if the contribution is a specific model and empirical evaluation, it may be necessary to either make it possible for others to replicate the model with the same dataset, or provide access to the model. In general. releasing code and data is often one good way to accomplish this, but reproducibility can also be provided via detailed instructions for how to replicate the results, access to a hosted model (e.g., in the case of a large language model), releasing of a model checkpoint, or other means that are appropriate to the research performed.
        \item While NeurIPS does not require releasing code, the conference does require all submissions to provide some reasonable avenue for reproducibility, which may depend on the nature of the contribution. For example
        \begin{enumerate}
            \item If the contribution is primarily a new algorithm, the paper should make it clear how to reproduce that algorithm.
            \item If the contribution is primarily a new model architecture, the paper should describe the architecture clearly and fully.
            \item If the contribution is a new model (e.g., a large language model), then there should either be a way to access this model for reproducing the results or a way to reproduce the model (e.g., with an open-source dataset or instructions for how to construct the dataset).
            \item We recognize that reproducibility may be tricky in some cases, in which case authors are welcome to describe the particular way they provide for reproducibility. In the case of closed-source models, it may be that access to the model is limited in some way (e.g., to registered users), but it should be possible for other researchers to have some path to reproducing or verifying the results.
        \end{enumerate}
    \end{itemize}

\item {\bf Open access to data and code}
    \item[] Question: Does the paper provide open access to the data and code, with sufficient instructions to faithfully reproduce the main experimental results, as described in supplemental material?
    \item[] Answer: \answerYes{} 
    \item[] Justification: We share code in a zip file. Note that it is not directly executable as users must directly apply to physionet to get access to the raw dataset and use another person's repo to load the data. 
    \item[] Guidelines:
    \begin{itemize}
        \item The answer NA means that paper does not include experiments requiring code.
        \item Please see the NeurIPS code and data submission guidelines (\url{https://nips.cc/public/guides/CodeSubmissionPolicy}) for more details.
        \item While we encourage the release of code and data, we understand that this might not be possible, so “No” is an acceptable answer. Papers cannot be rejected simply for not including code, unless this is central to the contribution (e.g., for a new open-source benchmark).
        \item The instructions should contain the exact command and environment needed to run to reproduce the results. See the NeurIPS code and data submission guidelines (\url{https://nips.cc/public/guides/CodeSubmissionPolicy}) for more details.
        \item The authors should provide instructions on data access and preparation, including how to access the raw data, preprocessed data, intermediate data, and generated data, etc.
        \item The authors should provide scripts to reproduce all experimental results for the new proposed method and baselines. If only a subset of experiments are reproducible, they should state which ones are omitted from the script and why.
        \item At submission time, to preserve anonymity, the authors should release anonymized versions (if applicable).
        \item Providing as much information as possible in supplemental material (appended to the paper) is recommended, but including URLs to data and code is permitted.
    \end{itemize}

\item {\bf Experimental Setting/Details}
    \item[] Question: Does the paper specify all the training and test details (e.g., data splits, hyperparameters, how they were chosen, type of optimizer, etc.) necessary to understand the results?
    \item[] Answer: \answerYes{}
    \item[] Justification: We share our training details in the Appendix, and we source our dataset split from a reproducibility study as highlighted in the text. 
    \item[] Guidelines:
    \begin{itemize}
        \item The answer NA means that the paper does not include experiments.
        \item The experimental setting should be presented in the core of the paper to a level of detail that is necessary to appreciate the results and make sense of them.
        \item The full details can be provided either with the code, in appendix, or as supplemental material.
    \end{itemize}

\item {\bf Experiment Statistical Significance}
    \item[] Question: Does the paper report error bars suitably and correctly defined or other appropriate information about the statistical significance of the experiments?
    \item[] Answer: \answerYes{} 
    \item[] Justification: We report 1-sigma error-bars where applicable. However, most of our human evaluation results have too small of a sample size to report additional error bars as we only had 2 human evaluators. 
    \item[] Guidelines:
    \begin{itemize}
        \item The answer NA means that the paper does not include experiments.
        \item The authors should answer "Yes" if the results are accompanied by error bars, confidence intervals, or statistical significance tests, at least for the experiments that support the main claims of the paper.
        \item The factors of variability that the error bars are capturing should be clearly stated (for example, train/test split, initialization, random drawing of some parameter, or overall run with given experimental conditions).
        \item The method for calculating the error bars should be explained (closed form formula, call to a library function, bootstrap, etc.)
        \item The assumptions made should be given (e.g., Normally distributed errors).
        \item It should be clear whether the error bar is the standard deviation or the standard error of the mean.
        \item It is OK to report 1-sigma error bars, but one should state it. The authors should preferably report a 2-sigma error bar than state that they have a 96\% CI, if the hypothesis of Normality of errors is not verified.
        \item For asymmetric distributions, the authors should be careful not to show in tables or figures symmetric error bars that would yield results that are out of range (e.g. negative error rates).
        \item If error bars are reported in tables or plots, The authors should explain in the text how they were calculated and reference the corresponding figures or tables in the text.
    \end{itemize}

\item {\bf Experiments Compute Resources}
    \item[] Question: For each experiment, does the paper provide sufficient information on the computer resources (type of compute workers, memory, time of execution) needed to reproduce the experiments?
    \item[] Answer: \answerYes{} 
    \item[] Justification: We have added a compute resources section in the Appendix.
    \item[] Guidelines:
    \begin{itemize}
        \item The answer NA means that the paper does not include experiments.
        \item The paper should indicate the type of compute workers CPU or GPU, internal cluster, or cloud provider, including relevant memory and storage.
        \item The paper should provide the amount of compute required for each of the individual experimental runs as well as estimate the total compute. 
        \item The paper should disclose whether the full research project required more compute than the experiments reported in the paper (e.g., preliminary or failed experiments that didn't make it into the paper). 
    \end{itemize}
    
\item {\bf Code Of Ethics}
    \item[] Question: Does the research conducted in the paper conform, in every respect, with the NeurIPS Code of Ethics \url{https://neurips.cc/public/EthicsGuidelines}?
    \item[] Answer: \answerYes{} 
    \item[] Justification: Our human evaluators are our collaborators and have undergone an internal process for ethical reporting.
    \item[] Guidelines:
    \begin{itemize}
        \item The answer NA means that the authors have not reviewed the NeurIPS Code of Ethics.
        \item If the authors answer No, they should explain the special circumstances that require a deviation from the Code of Ethics.
        \item The authors should make sure to preserve anonymity (e.g., if there is a special consideration due to laws or regulations in their jurisdiction).
    \end{itemize}

\item {\bf Broader Impacts}
    \item[] Question: Does the paper discuss both potential positive societal impacts and negative societal impacts of the work performed?
    \item[] Answer: \answerNA{}{} 
    \item[] Justification: We felt much of our work was too preliminary to have direct consequences. However, we feel that improving the interpretability of these models will generally be positive as its research is directly tied to safety rather than extraneous intents.
    \item[] Guidelines:
    \begin{itemize}
        \item The answer NA means that there is no societal impact of the work performed.
        \item If the authors answer NA or No, they should explain why their work has no societal impact or why the paper does not address societal impact.
        \item Examples of negative societal impacts include potential malicious or unintended uses (e.g., disinformation, generating fake profiles, surveillance), fairness considerations (e.g., deployment of technologies that could make decisions that unfairly impact specific groups), privacy considerations, and security considerations.
        \item The conference expects that many papers will be foundational research and not tied to particular applications, let alone deployments. However, if there is a direct path to any negative applications, the authors should point it out. For example, it is legitimate to point out that an improvement in the quality of generative models could be used to generate deepfakes for disinformation. On the other hand, it is not needed to point out that a generic algorithm for optimizing neural networks could enable people to train models that generate Deepfakes faster.
        \item The authors should consider possible harms that could arise when the technology is being used as intended and functioning correctly, harms that could arise when the technology is being used as intended but gives incorrect results, and harms following from (intentional or unintentional) misuse of the technology.
        \item If there are negative societal impacts, the authors could also discuss possible mitigation strategies (e.g., gated release of models, providing defenses in addition to attacks, mechanisms for monitoring misuse, mechanisms to monitor how a system learns from feedback over time, improving the efficiency and accessibility of ML).
    \end{itemize}
    
\item {\bf Safeguards}
    \item[] Question: Does the paper describe safeguards that have been put in place for responsible release of data or models that have a high risk for misuse (e.g., pretrained language models, image generators, or scraped datasets)?
    \item[] Answer: \answerNA{}{} 
    \item[] Justification: Our methods are for interpretability rather than generative purposes.
    \item[] Guidelines:
    \begin{itemize}
        \item The answer NA means that the paper poses no such risks.
        \item Released models that have a high risk for misuse or dual-use should be released with necessary safeguards to allow for controlled use of the model, for example by requiring that users adhere to usage guidelines or restrictions to access the model or implementing safety filters. 
        \item Datasets that have been scraped from the Internet could pose safety risks. The authors should describe how they avoided releasing unsafe images.
        \item We recognize that providing effective safeguards is challenging, and many papers do not require this, but we encourage authors to take this into account and make a best faith effort.
    \end{itemize}

\item {\bf Licenses for existing assets}
    \item[] Question: Are the creators or original owners of assets (e.g., code, data, models), used in the paper, properly credited and are the license and terms of use explicitly mentioned and properly respected?
    \item[] Answer: \answerYes{} 
    \item[] Justification: We cite many of the baselines that are used in this paper.
    \item[] Guidelines:
    \begin{itemize}
        \item The answer NA means that the paper does not use existing assets.
        \item The authors should cite the original paper that produced the code package or dataset.
        \item The authors should state which version of the asset is used and, if possible, include a URL.
        \item The name of the license (e.g., CC-BY 4.0) should be included for each asset.
        \item For scraped data from a particular source (e.g., website), the copyright and terms of service of that source should be provided.
        \item If assets are released, the license, copyright information, and terms of use in the package should be provided. For popular datasets, \url{paperswithcode.com/datasets} has curated licenses for some datasets. Their licensing guide can help determine the license of a dataset.
        \item For existing datasets that are re-packaged, both the original license and the license of the derived asset (if it has changed) should be provided.
        \item If this information is not available online, the authors are encouraged to reach out to the asset's creators.
    \end{itemize}

\item {\bf New Assets}
    \item[] Question: Are new assets introduced in the paper well documented and is the documentation provided alongside the assets?
    \item[] Answer: \answerYes{} 
    \item[] Justification: We note that we will be submitting code with our submission, anonymized with a brief readme file highlighting where to look for the data required to run the code. However, we note much of the code should be pretty straightforward and are at least somewhat documented.
    \item[] Guidelines:
    \begin{itemize}
        \item The answer NA means that the paper does not release new assets.
        \item Researchers should communicate the details of the dataset/code/model as part of their submissions via structured templates. This includes details about training, license, limitations, etc. 
        \item The paper should discuss whether and how consent was obtained from people whose asset is used.
        \item At submission time, remember to anonymize your assets (if applicable). You can either create an anonymized URL or include an anonymized zip file.
    \end{itemize}

\item {\bf Crowdsourcing and Research with Human Subjects}
    \item[] Question: For crowdsourcing experiments and research with human subjects, does the paper include the full text of instructions given to participants and screenshots, if applicable, as well as details about compensation (if any)? 
    \item[] Answer: \answerYes{} 
    \item[] Justification: We share the prompts we give our human annotators and LLMs in the Appendix.
    \item[] Guidelines:
    \begin{itemize}
        \item The answer NA means that the paper does not involve crowdsourcing nor research with human subjects.
        \item Including this information in the supplemental material is fine, but if the main contribution of the paper involves human subjects, then as much detail as possible should be included in the main paper. 
        \item According to the NeurIPS Code of Ethics, workers involved in data collection, curation, or other labor should be paid at least the minimum wage in the country of the data collector. 
    \end{itemize}

\item {\bf Institutional Review Board (IRB) Approvals or Equivalent for Research with Human Subjects}
    \item[] Question: Does the paper describe potential risks incurred by study participants, whether such risks were disclosed to the subjects, and whether Institutional Review Board (IRB) approvals (or an equivalent approval/review based on the requirements of your country or institution) were obtained?
    \item[] Answer: \answerYes{} 
    \item[] Justification: Since there are no risks in this human study with our collaborators, we note that there was no need for an IRB study as we are not crowd sourcing anything. We have made a note of this in the Appendix.
    \item[] Guidelines:
    \begin{itemize}
        \item The answer NA means that the paper does not involve crowdsourcing nor research with human subjects.
        \item Depending on the country in which research is conducted, IRB approval (or equivalent) may be required for any human subjects research. If you obtained IRB approval, you should clearly state this in the paper. 
        \item We recognize that the procedures for this may vary significantly between institutions and locations, and we expect authors to adhere to the NeurIPS Code of Ethics and the guidelines for their institution. 
        \item For initial submissions, do not include any information that would break anonymity (if applicable), such as the institution conducting the review.
    \end{itemize}

\end{enumerate}

\end{document}

%% file: sections/1introduction.tex
\section{Introduction} 
\begin{wrapfigure}{r}{0.3\textwidth}
    \centering
    \includegraphics[width=0.3\textwidth]{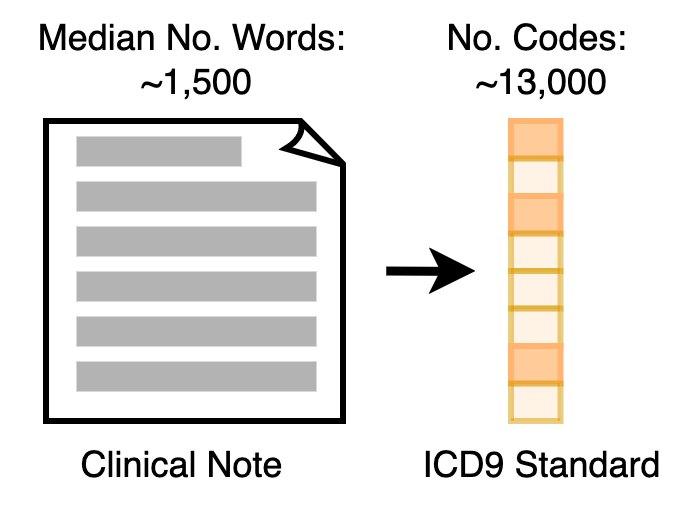}
    \caption{Medical coding as a high dimensional multilabel prediction task.}
    \label{fig:MedicalCodingExample}
\end{wrapfigure}
High-dimensional or extreme multilabel prediction, where potentially thousands of non-exclusive labels are predicted from high-dimensional inputs, poses significant challenges in interpretability in high-stakes domains such as healthcare. Due to the extreme range of potential inputs and outputs in these contexts, expert annotations are often scarce and costly, making the evaluation of interpretable multilabel models difficult. One prime example is medical coding, which involves assigning potentially tens of thousands of International Classification of Diseases (ICD) codes to lengthy, unstructured clinical notes for categorizing diagnoses and procedures \cite{HistoryOfICD}. This complex and time-consuming task requires explicit medical expertise, making human annotation expensive \cite{OMalley2005-we-icd-coding-hard}. While recent advancements in pre-trained language models have significantly improved coding efficiency and accuracy by treating it as a high-dimensional multilabel classification problem \cite{AutomatedMedicalCoding, huang2022plmicd, YAN2022161SurveyICDCoding}, the opaque nature of these black-box models raises concerns about their decision-making processes and potential biases \cite{HAKKOUM2022108391_medical_interpretability_review, räuker2023transparent_survey_mechanistic}. To uphold transparency and maintain patient trust, these models must be capable of explaining their code predictions, which are crucial for billing, research, and clinical treatment purposes \cite{Rao2022-kj_TransparencyMedicalCare, Johnson2021-aw-ICD-coding-injury}. This is especially the case where misclassifications can directly impact patient outcomes, underscoring the need for interpretable and transparent AI models \cite{Johnson2021-aw-ICD-coding-injury}.

 Despite significant progress in the post-hoc interpretation of black-box models, several key issues remain in leveraging them for extreme multilabel prediction, such as medical coding. Model agnostic attribution methods, such as SHAP \cite{lundberg2017unifiedshap,chen2021interpretation_multilabel}, which measure the change in a model's output given a smart perturbation in the input space, are computationally infeasible for high-dimensional problems where clinical notes consist of thousands of tokens with an equally large multilabel prediction space  \cite{Lundberg2020ShapLocalStuffForTrees, Chen2022SeriesOfModels, shrikumar2019learningDeepLift, mosca-etal-2022-shap-NLP}. Post-hoc mechanistic interpretability relies on elucidating the functions of neuron subsets like the label attention mechanism, mapping each token to an ICD code, a common approach in medical coding \cite{chaudhari2021attentionSurvey, vaswani2023attentionIsAllYouNeed, Vu_2020_LAAT, YAN2022161SurveyICDCoding, mrini2020rethinking_og_label_attn}. However, due to the nonlinear projection underlying the construction of the label attention matrix, this mechanism is largely uninterpretable, hiding the learned relationships required to generate each attribution. Furthermore, feature attribution maps generated by such attention mechanisms are often incomplete explanations where its faithfulness to the prediction can be questionable and often fail to provide insights into model behavior beyond attribution \cite{serrano2019attentionnotInterpretable, Zhang_2021SurveyNNInterpretability, räuker2023transparent_survey_mechanistic}. In summary, these interpretability methods are generally limited to local interpretability, only capable of explaining predictions on a per-example basis \cite{Survey_Machine_Learning}.
 
 Circuit analysis uses causal frameworks to identify interpretable circuits within dense language models~\cite{conmy2023automated_circuit, räuker2023transparent_survey_mechanistic}. However, most existing work has focused on simple tasks where counterfactuals exist and the problem's dimensionality is relatively low  \cite{conmy2023automated_circuit}. For problems with larger input and output spaces, such as high-dimensional multilabel ICD coding, these methods are often computationally infeasible, and approximation methods are imperfect even for simple tasks \cite{kramár2024atp}. Thus, a persistent gap exists between accurate model explanations and computational constraints for high dimensional multilabel predictions.

To bridge this gap, intrinsically explainable models have been developed. For instance, prototype models generate explanations from neural layers where input samples are directly compared to prototypical examples of each label~\cite{tang2023protoeegnet, ma2023lookslikethose}. While effective, this approach requires human annotators to design a sufficiently diverse corpus of exemplary examples for each class. In the ICD coding space\cite{AutomatedMedicalCoding}, there are privacy concerns related to using real patient data to construct such prototypes. Additionally, in terms of scalability, there are tens of thousands of ICD codes with an equally large number of diverse examples of clinical notes and settings \cite{HistoryOfICD,YAN2022161SurveyICDCoding}, making the construction of a diverse prototype corpus infeasible. 

Finally, white-box approaches have been attempted using sparsity, where transformer models are regularized during training to transform data into a mathematically interpretable form \cite{yu2023whiteboxcrate}, effectively mitigating polysemanticity or superposition seen in neurons \cite{elhage2022superposition}. However, pre-training new models can be costly.  To attain similar levels of performance, more model parameters are needed than their dense counterparts \cite{yu2023whiteboxcrate}. Fortunately, studies have shown that increasing sparsity in decision-related layers significantly enhances the overall interpretability of the prediction process by demonstrating that sparsity results in neurons activating only for specific data features \cite{wong2021leveraging_debuggable_networks, thompson2024contextual_sparse_interpretability}.

 Inspired by these results, we propose an interpretable DIctionary Label Attention (\method) module incorporating sparsity via dictionary learning \cite{OLSHAUSEN19973311_OGSparseCoding, bricken2023monosemanticity}. Additionally, to enhance the interpretability of the learned sparse dictionary features without needing expert annotations,  we leverage medical large language models (LLMs) to automatically interpret our learned sparse abstractions. 

We evaluate three aspects of our approach \method and show: 
\begin{itemize}[leftmargin=*]
\item Interpretability: The dictionary label attention layer in \method is more human-interpretable than its dense counterparts. Learning sparse medical abstractions enables insights into the model's decision-making process for high-dimensional multilabel predictions, addressing the challenges of interpretability in complex domain-specific scenarios.
\item Scalability: While imperfect, we demonstrate that medical LLMs can serve as capable domain-specific dictionary feature summarizers and annotators. This enables the scalability of deep auto-interpretability pipelines, overcoming the limitations of manual annotation and prototype design in the medical domain, where expert knowledge is scarce and expensive.
\item Performance: Despite the incorporation of sparsity and interpretability, \method maintains competitive performance with current state-of-the-art black-box baselines on the large cleaned MIMIC-III dataset \cite{AutomatedMedicalCoding}, demonstrating its effectiveness in extreme multilabel prediction tasks like medical coding.
\end{itemize}




%% file: sections/2related_work.tex
\vspace{-0.2cm} 
\section{Related Work}
\vspace{-0.1cm} 
\subsection{Dictionary Learning}
Much of our work was directly inspired by the use of dictionary learning to better understand the predictions made by auto-regressive LLMs \cite{bricken2023monosemanticity, cunningham2023sparse, yun2023transformer-residual} in a post-hoc manner. However, we note that the sparse coding problem \cite{Zhang_2015_survey_sparse, OLSHAUSEN19973311_OGSparseCoding} is not unique to the language domain and has been applied to improve interpretability in various other modalities such as vision \cite{yu2023whiteboxcrate, ghosh2023dictionary_representations, learning_dictionary_for_visual_attention} and time-series \cite{xu2023_dictionary_time_series, tang2023dictioanry_trajectories} tasks. To our knowledge, such an approach has not been directly leveraged to explain deep extreme multi-label prediction settings better, as seen in the automated ICD coding task where the input space consists of thousands of tokens with an equally large prediction space.

\vspace{-0.2cm} 
\subsection{Interpretability in Automated ICD Coding}
There exists a plethora of attempts in making interpretable automated ICD coding methods \cite{YAN2022161SurveyICDCoding}. For instance, phrase matching \cite{cao-etal-2020-clinical_ICD_phrase_matching} and phrase extraction using manually curated knowledge bases \cite{DUQUE2021102177_Knowledge_base}, offer inherent interpretability but fall short in expressive power compared to neural network-based approaches.  

On the other hand, the prevailing interpretability method for deep neural models in ICD coding tasks rely on the label attention (LAAT) mechanism \cite{YAN2022161SurveyICDCoding}. Specifically, the LAAT mechanism projects token embeddings into a label-specific attention space, where each token receives a score indicating its relevance to each ICD prediction. Such attention-based associations between tokens and classes have been employed in various architectures, including convolutional models like CAML \cite{mullenbach2018explainable_CAML}, recurrent neural networks \cite{Vu_2020_LAAT}, and pre-trained language models \cite{huang2022plmicd}. While computationally efficient in highlighting tokens locally relevant to each ICD prediction, its nonlinear projections prevent direct interpretation of the mechanisms behind each attribution score. Furthermore, its dense pretrained language model (PLM) embeddings are difficult to interpret due to polysemanticity \cite{olah2020zoomPolySemanticity}, hindering direct understanding of the global mechanisms behind each ICD code prediction. Other attempts have been made through prompting large language models (LLMs) \cite{yang2023surpassing_GPT4_LLM_interp} where the LLM retrieves the evidence behind each prediction. However, LLMs are known to hallucinate \cite{zhao2023surveyofLLMs,huang2023surveyhallucination,huang2023largeLMexplainthemselves}, reducing the faithfulness of their explanations. This discrepancy highlights a persistent tradeoff between interpretability and performance in ICD coding tasks.

Our method \method, in comparison, directly interprets and disentangles concepts learned within the PLM embedding space. It maps global medically relevant concepts in the form of dictionary features to each ICD code while retaining the local interpretability derived from the label attention mechanism.

%% file: sections/3method.tex
\vspace{-0.4cm} 
\section{DILA} \label{sec : DILA}
\begin{figure*}[h] 
\centering
\includegraphics[width=1.0\textwidth]{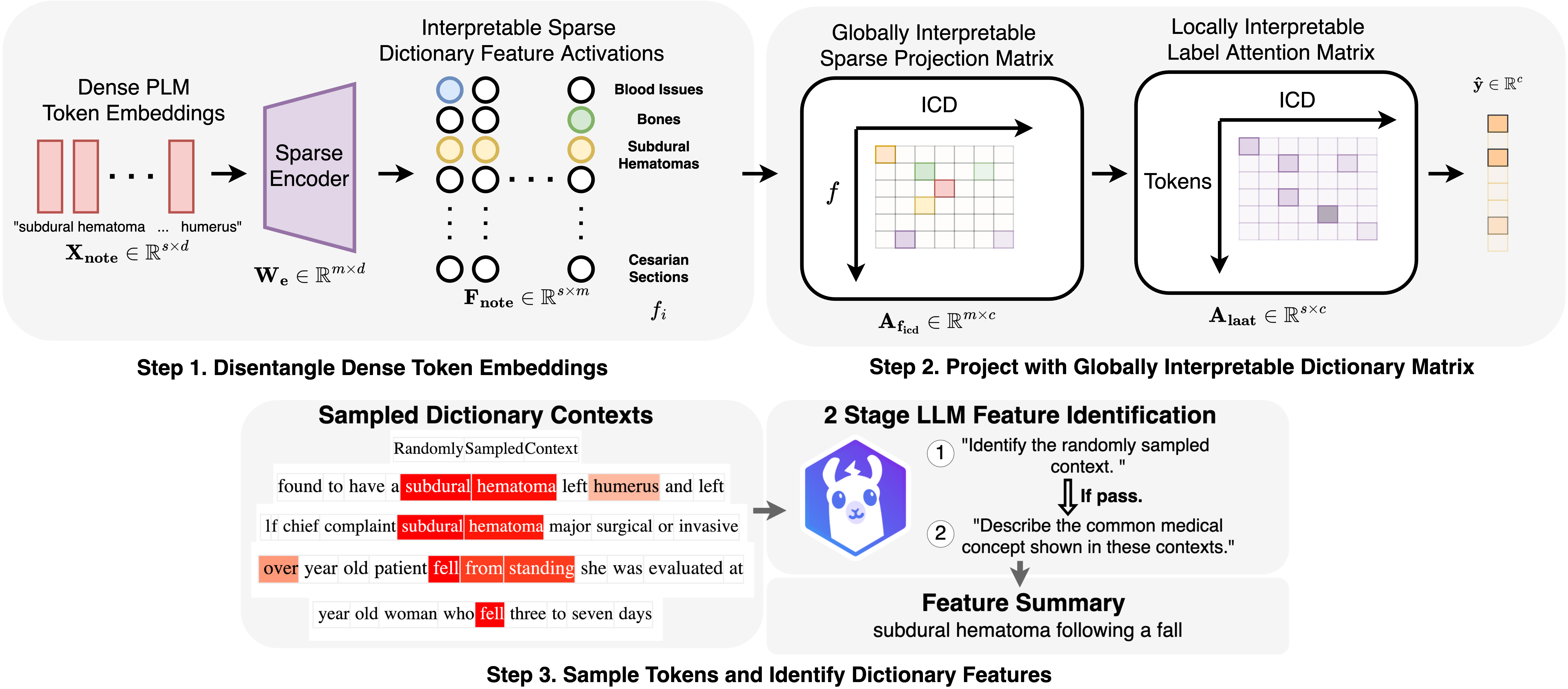}
\caption{DILA composes of three steps: First, we disentangle each token embedding into its dictionary features. Then, we project each set of dictionary features with our globally interpretable $\mathbf{A_{f_\text{icd}}}$ to generate our local explanation $\mathbf{A_{\text{laat}}}$ for downstream multilabel prediction. Finally, medical LLMs identify the learned dictionary feature to understand the learned relationships in $\mathbf{A_{f_\text{icd}}}$. }
\label{fig:dla_overview}
\end{figure*}

\textbf{Overview.} DILA consists of three key components: (1) dictionary learning to disentangle dense embeddings into sparse, interpretable dictionary features; (2) a dictionary label attention module that utilizes a sparse, interpretable matrix capturing the global relationships between dictionary features and medical codes to generate each clinical note's label attention matrix, representing the local token-code relationships; and (3) an automated interpretability approach using medical LLMs, as shown in Figure \ref{fig:dla_overview}.

\vspace{-0.2cm} 
\subsection{Dictionary Learning} \label{sec: DL}
Dictionary learning decomposes vectors $\mathbf{x}$ that often represent tokens or words into a sparse linear combination of basis vectors. We formulate this problem using a sparse autoencoder with dense PLM embeddings $\mathbf{x} \in \mathbb{R}^d$, sparse dictionary feature activations $\mathbf{f} \in \mathbb{R}^m$, as well as encoder and decoder weight matrices $\mathbf{W_e} \in \mathbb{R}^{d \times m}$ and $ \mathbf{W_d} \in \mathbb{R}^{m \times d}$ with corresponding bias units $\mathbf{b_e} \in \mathbb{R}^m, \mathbf{b_d} \in \mathbb{R}^d$:

\begin{align}
\bar{\mathbf{x}} &= \mathbf{x} - \mathbf{b_d} \label{eq:L1 Autoencoder} \\
\mathbf{f} &= \text{ReLU}(\mathbf{W_e} \bar{\mathbf{x}} + \mathbf{b_e}) \\
\hat{\mathbf{x}} &= \mathbf{W_d} \mathbf{f} + \mathbf{b_d} \\
\mathcal{L}_{\text{saenc}} &= \frac{1}{|X|} \sum_{\mathbf{x} \in X} \left\lVert \mathbf{x} - \hat{\mathbf{x}} \right\rVert_2^2 + \lambda_{L_1} \left\lVert \mathbf{f} \right\rVert_1 + \lambda_{L_2} \left\lVert \mathbf{f} \right\rVert_2^2 \label{eq: L1 Loss}
\end{align}

The $\lambda_{L_1} \left\lVert \mathbf{f} \right\rVert_1$ and $\lambda_{L_2}  \left\lVert \mathbf{f} \right\rVert_2^2$ elastic loss terms \cite{elastic-net} in Equation \eqref{eq: L1 Loss} enforce sparsity on the encoded dictionary feature activations $\mathbf{f}$, 
ensuring that only specific $f_i \in \mathbf{f}$ activate (i.e., are nonzero) for specific token embeddings $\mathbf{x}$. This sparsity influences each element $f_i$ to be more interpretable, often corresponding to a specific medical concept. To ground these interpretations within the embedding space, we reconstruct $\mathbf{x}$ using our dictionary embeddings $\mathbf{W_d} = \begin{bmatrix} \mathbf{h_{0}}, & \mathbf{h_{1}}, & \ldots &, \mathbf{h_{m}} \end{bmatrix}^T$, allowing any $\mathbf{x}$ to be represented through a sparse linear combination of $\mathbf{h_i}$ as defined by our sparse autoencoder: $\mathbf{x} \approx \sum_i^m f_i  \mathbf{h_{i}}$.

To construct a human interpretable dictionary, we sample token embeddings from a text corpus and sort for the highest activating tokens for $f_i$, called a dictionary feature's context, as they describe the underlying meaning of each $f_i$ (see Appendix \ref{Appendix: Dictionary Construction}). For instance, in step 3 of Figure \ref{fig:dla_overview}, the contexts of a dictionary feature $f_i$ all share the same theme of falls and subdural hematomas, a type of head injury.

\vspace{-0.2cm} 
\subsection{Dictionary Label Attention} \label{sec : DLA}
We propose a simpler, disentangled version of label attention called dictionary label attention that maps the concepts represented by each of the $m$ dictionary features $f_i$ to their corresponding ICD codes (Figure \ref{fig:dla_overview}). Given a clinical note's tokenized PLM embeddings $\mathbf{X_{\text{note}}} \in \mathbb{R}^{s \times d}$ of length $s$, we encode them into disentangled dictionary features $\mathbf{F_{\text{note}}} \in \mathbb{R}^{s \times m}$ using a sparse autoencoder. We initialize the sparse projection matrix $\mathbf{A_{f_\text{icd}}} \in \mathbb{R}^{m \times c}$, which maps the relationship between each of the $m$ dictionary features $f_i$ and $c$ ICD codes, by encoding the tokens in each ICD code's description into their respective dictionary features $\mathbf{F}_{\text{desc}}^{(c)} \in \mathbb{R}^{l \times m}$, where $l$ is the description length. We then average pool these features into $\mathbf{f^{(c)}} \in \mathbb{R}^{m}$ and perform the following operations:
\begin{align}
\mathbf{A_{f_\text{icd}}} &= \begin{bmatrix} \mathbf{f}^{(1)}, & \mathbf{f}^{(2)}, & \ldots &, \mathbf{f}^{(c)} \end{bmatrix} \in \mathbb{R}^{m \times c} \label{eq: initialization of Aficd} \\
\mathbf{A_{\text{laat}}} &= \text{softmax}(\mathbf{F}_{\text{note}} \mathbf{A_{f_\text{icd}}}) \in \mathbb{R}^{s \times c} \\
\mathbf{X_{\text{att}}} &= \mathbf{A_{\text{laat}}}^T \mathbf{X}_{\text{note}} \in \mathbb{R}^{c \times d} 
\end{align}
Once the label aware representation $\mathbf{X_{\text{att}}}$ is computed, it is passed through a decision layer for the final prediction $\mathbf{\hat{y}} \in \mathbb{R}^c$.

\textbf{Relationship to Dense Label Attention.} In contrast to the original label attention mechanism in \cite{Vu_2020_LAAT}, we have essentially replaced the original nonlinear projection of $\mathbf{A_{\text{laat}}} = \text{softmax}(\mathbf{Z} \mathbf{W_c})$ where $\mathbf{Z} = \tanh{(\mathbf{X_{\text{note}}} \mathbf{W_z})}$ with a single linear sparse projection matrix $\mathbf{A}_{f_\text{icd}}$, representing each of the $c$ ICD codes as a set of dictionary features. Unlike $\mathbf{Z}$, since $\mathbf{f}$ is always positive, every element's magnitude in $\mathbf{A}_{f_\text{icd}}$ indicates the strength of the overall relationship between a dictionary feature $f_i$ and an ICD code $\mathbf{\hat{y}_i}$. 

\textbf{Training.} Training consists of two steps. First, we train a sparse autoencoder on all of the embeddings generated by our PLM within the training set. Then, we initialize our label attention module, and do end-to-end training using a combination of our sparse autoencoder loss defined in equation \ref{eq: L1 Loss} and the binary cross entropy loss function. We use an additional hyperparameter $\lambda_{\text{saenc}}$ to prevent $L_{\text{saenc}}$ from dominating $L_{\text{BCE}}$, giving us our final loss function in equation \ref{eq: Overall Loss Function}. 
\begin{equation} \label{eq: Overall Loss Function}
    L = \lambda_{\text{saenc}} L_{\text{saenc}} + L_{\text{BCE}}
\end{equation}
\vspace{-0.2cm} 
\subsection{Auto-Interpretability} \label{sec: autointerp}
One major challenge in our framework is that the sampled dictionary contexts and its sparse dictionary feature representations $\mathbf{f}$ are unlabeled. Ideally, human expert annotators \cite{bricken2023monosemanticity} would inspect the dictionary features' highest activating contexts. However, unlike general language disciplines \cite{zhang2023needle_mturk}, expert medical professional annotators are not as readily available and often lack the time to provide high-quality annotations for each learned dictionary feature. An automated pipeline to identify and quantify the quality of such dictionaries is crucial for scalable interpretability, as medical PLMs learn thousands of domain-specific concepts. Inspired by \cite{subramanian2017spine}'s feature identification and \cite{bills2023language_openai_autointerpret}'s LLM neuron interpretability experiments, we develop a two-stage auto-interpretability pipeline shown in Figure \ref{fig:dla_overview}. 

More specifically, to discern whether or not the LLM can identify the dictionary feature faithfully, we perform an identification test that asks the LLM to identify the randomly sampled context that does not activate the dictionary feature representing a specific medical concept out of five total contexts. If the LLM is capable of discerning the outlier context, it implies it can understand that the other contexts all belong to the same underlying medical concept. In practice, we prompt it with only the highlighted tokens (red) to avoid information contamination from the context window as there can often be neighboring medical concepts that may misdirect the LLM. Once determined to be identifiable, we prompt the LLM to summarize the dictionary feature given the original contexts.



%% file: sections/4results_discussion.tex
\vspace{-0.2cm} 
\section{Experiments}
\textbf{Dataset.}  We use a similar medical coding dataset as \cite{mullenbach2018explainable_CAML}. Specifically, we train and evaluate our model on the revised "MIMIC-III clean" data split sourced from a reproducibility study \cite{AutomatedMedicalCoding}.

\textbf{Models.} We mainly explore two models, our method \method, and its nearest dense label attention comparison PLM-ICD \cite{huang2022plmicd}, both of which use the same text medical RoBERTa PLM. While we reuse the weights provided by \cite{AutomatedMedicalCoding} in our interpretability comparison, we also retrain their model using the same training hyperparameters used in \method to get a more direct comparison in performance in Section \ref{sec: perf} as we were unable to replicate the reported large batch size of 16 \cite{AutomatedMedicalCoding} due to GPU limitations. We report the training hyperparameters used in the Appendix (\ref{Appendix : Training Details}).

\textbf{Annotators.} We ask two medical experts for our human evaluations, a clinically licensed physician and a medical scientist trainee with extensive clinical training. For the LLM, we use a state-of-the-art quantized, medically fine-tuned Llama 3 OpenBioLLM 70b model in our auto-interpretability pipeline.

\textbf{Overview.} We explore the interpretability, mechanistic insights, and performance of our proposed method, \method, for automated ICD coding. Using human evaluations, we assess the human understandability of its learned dictionary features and the efficacy of the auto-interpretability pipeline, and demonstrate its ability to efficiently provide precise global explanations through ablation studies, visualizations, and human predictability experiments. Finally, we evaluate \method's performance compared to baselines. 
\vspace{-0.2cm} 
\subsection{Automated Interpretability of Dictionary Features} \label{results: auto interp}
\textbf{Setup.} In Figure \ref{fig:dla_overview}, the interpretation of the global projection matrix from step 2 requires identifying the underlying abstractions learned for each dictionary feature $f_i$ with the auto-interpretability pipeline. There are two main components to the auto-interpretability pipeline of \method, 1) the identification of dictionary features, and 2) the summarization of these dictionary features given their sampled context tokens from step 3 in Figure \ref{fig:dla_overview}. 

To evaluate our dictionary feature identification method and quantify the interpretability or human understandability of our trained dictionaries, we conduct a medical expert identification experiment across 100 randomly sampled dictionary features, as described in Section \ref{sec: autointerp}. An example of the LLM identification prompt is showcased in the Appendix (\ref{appendix: LLM identfication}). 
 
As hallucination is always plausible, human medical experts are asked to evaluate the summaries generated by the LLMs of the identified dictionary features. We ask whether they agree with the originally sampled contexts and how confident they are in their responses, from 1, being unsure, to 4, absolute confidence.

\textbf{Baselines.} Furthermore, we compare our dictionary features to dense $\mathbf{Z}$ activations from the label attention mechanism (Section \ref{sec : DLA}) in PLM-ICD \cite{huang2022plmicd} and a random baseline where contexts are randomly sampled from a large medical token corpus from the test set. We run our automated pipeline across all observed features: 6,088 active features in our dictionary $\mathbf{f}$, 768 features in our dense $\mathbf{Z}$, and 1,000 randomly sampled contexts. However, since we felt random context summaries were trivial, we only ask our annotators to evaluate the summaries generated from interpreting the dictionary features $\mathbf{f}$ and its dense label attention counterpart $\mathbf{Z}$.

\begin{table}[h]
\centering
\resizebox{\textwidth}{!}{
\begin{tabular}{l*{7}{c}}
\toprule
& \makecell{Medical Expert 1\\(100) $\uparrow$} & \makecell{Medical Expert 2\\(100) $\uparrow$} & \makecell{LLM\\(100) $\uparrow$} & \makecell{LLM\\(all) $\uparrow$} & \makecell{Avg. Cosine\\Similarity $\uparrow$} & \makecell{Avg. Jaccard\\Similarity $\uparrow$} & \makecell{No. of LLM\\Identified Features $\uparrow$}\\
\midrule
Dict. $\mathbf{f}$ (\method) & \textbf{0.67} \textbf{(+55.8\%)} & \textbf{0.69} \textbf{(+64.3\%)} & \textbf{0.59} \textbf{(+73.5\%)} & \textbf{0.58} \textbf{(+70.6\%)} & \textbf{0.77} \textbf{(+30.5\%)} & \textbf{0.62} \textbf{(+44.2\%)} & \textbf{3,524} \textbf{(+1239.9\%)} \\
Dense $\mathbf{Z}$ & 0.43 & 0.42 & 0.34 & 0.34 & 0.59 & 0.43 & 263 \\
Random & 0.23 & 0.19 & 0.27 & 0.19 & 0.15 & 0.08 & 193 \\
\bottomrule
\end{tabular}
}
\caption{Identification test accuracy comparing human experts and LLMs on interpreting dictionary features ($\mathbf{f}$), dense embeddings ($\mathbf{Z}$), and random contexts. The results show the superior performance of human experts and the importance of sparse embeddings for interpretability. The relative improvement of Dict. $\mathbf{f}$ (\method) over Dense $\mathbf{Z}$ is shown in parentheses.}
\label{tab:identification test}
\end{table}
\vspace{-0.2cm} 
\textbf{Identification of Dictionary Features.} Table \ref{tab:identification test} demonstrates improved interpretability by leveraging a sparse embedding $\mathbf{f}$ over the dense embedding $\mathbf{Z}$, with significantly more interpretable $\mathbf{f}$ features (3,524) compared to $\mathbf{Z}$ features (263), suggesting over 3,000 interpretable abstract concepts were hidden in superposition \cite{elhage2022superposition}. Comparing LLM and human responses using a vector similarity metric reveals that human expert annotations identify more interpretable features and that human alignment of medical LLM feature identifications deteriorates as interpretability declines, suggesting a substantial gap between domain-specific annotators and medical LLMs. Further qualitative examinations reveal that LLMs fail to identify features with intrinsic relationships between contexts not obvious by language, such as linking "banding" and blood loss in Appendix Figure \ref{fig: f_human_ID}. However, given the volume of identified interpretable features and the lack of extensive prompt engineering, the potential to use LLMs for automatically interpreting previously black-box models is significant, dramatically reducing the number of human annotations needed for obviously interpretable features.

\begin{table}[h]
\centering
\resizebox{0.8\textwidth}{!}{
\begin{tabular}{lcccc}
\hline
& Expert 1, Agreement $\uparrow$ & Expert 2, Agreement$\uparrow$ & Expert 1, Confidence $\uparrow$ & Expert 2, Confidence$\uparrow$ \\
\hline
Dict. $\mathbf{f}$ (\method) & $\mathbf{0.83}$ \textbf{(+5.1\%)} & $\mathbf{0.92}$ \textbf{(+12\%)} & $\mathbf{3.85} \pm \mathbf{0.41}$  & $\mathbf{3.80} \pm \mathbf{0.45}$ \\
Dense $\mathbf{Z}$ & $0.79$ & $0.82$ & $3.79 \pm 0.41$ & $3.44 \pm 0.75$ \\
\hline
\end{tabular}}
\caption{Human evaluations of LLM summaries of dictionary features. We report the standard deviations of our confidence scores. The relative percentage improvement in LLM summary agreement of Dict. $\mathbf{f}$ (\method) over Dense $\mathbf{Z}$ are shown in parentheses.}
\label{tab:summary agreement}
\end{table}
\vspace{-0.2cm} 
\textbf{Quality of LLM Dictionary Feature Summarizations.} From Table \ref{tab:summary agreement}, we see that the majority of the summaries generated by the LLMs are in agreement with our human evaluators. Our qualitative evaluations show that the contexts of summaries rejected by our human annotators for the dictionary features are substantially more coherent than the ones rejected by the ones in the dense neurons $\mathbf{Z}$ as shown in Appendix Section \ref{Appendix: Human Summary Eval}. Crucially, many of the dictionary feature summaries that were rejected, were rejected due to their lack of specificity rather than being unrelated to the dictionary contexts or hallucinations, allowing us to better conduct mechanistic interpretability experiments.

\vspace{-0.2cm} 
\subsection{Mechanistic Interpretability} 
\textbf{Runtime Performance.} One crucial utility of mechanistic interpretability is the ability to generate efficient interpretations of a model's predictions, whether locally or globally. In Table \ref{tab:Run Times}, we show that using an off-the-shelf KernelSHAP \cite{lundberg2017unifiedshap} interpreter to analyze the dense PLM embeddings of a single clinical note is extremely computationally expensive, when compared to its mechanistic counterparts, highlighting its impractical use in high dimensional multilabel prediction. 
\begin{table}[h]
\centering
\resizebox{0.7\textwidth}{!}{
\begin{tabular}{lccccc}
\toprule
& \multicolumn{1}{c}{Model-agnostic} & \multicolumn{3}{c}{Mechanistic Interpretability (\method)} \\
\cmidrule(lr){2-2} \cmidrule(lr){3-5}
& KernelSHAP \cite{lundberg2017unifiedshap} & Computing $\mathbf{A_{\text{laat}}}$ & Encoding $\mathbf{F_{\text{note}}}$ & Accessing $\mathbf{A_{f_\text{icd}}}$ \\
\midrule
Time $\downarrow$ & 62m 10.71s & 0.04s & 0.03s & 0.00s \\
\bottomrule
\end{tabular}}
\captionsetup{name=Table}
\caption{Runtimes for interpreting a single clinical note using different methods. KernelSHAP, a model-agnostic method, is used to interpret the high-dimensional token embeddings generated by the PLM. The other runtimes are from our DILA method, demonstrating the efficiency of mechanistic interpretability compared to black-box approaches. Accessing the globally interpretable matrix $\mathbf{A_{f_\text{icd}}}$ is virtually instantaneous.}
\label{tab:Run Times}
\end{table}

\textbf{Setup.} To evaluate the explainability of our $\mathbf{A_{f_\text{icd}}}$ matrix, we compare to the common local attribution approach of token-based attention ($\mathbf{A_{\text{laat}}}$) through an ablation study. For each clinical note, we identify the highest softmax probability ICD code and ablate the weights corresponding to the observed activated dictionary features $f_i$, and measure the softmax probability drop for the target code and the sum of absolute changes for other codes.

\textbf{Baselines.} To compare against alternative local interpretability approaches, we identify the most relevant tokens for the highest-probability ICD code using $\mathbf{A_{\text{laat}}}$ and perturb the PLM embeddings by ablating, noising, or replacing these tokens with medically irrelevant ones, measuring the same downstream effects.

\begin{table}[h]
    \centering
    \resizebox{0.8\textwidth}{!}{
    \begin{tabular}{lcccc}
    \toprule
    & $\mathbf{A_{f_\text{icd}}}$ Weight Ablation & $\mathbf{A_{\text{laat}}}$ Token Ablation & $\mathbf{A_{\text{laat}}}$ Token Noising & $\mathbf{A_{\text{laat}}}$ Token Replacement \\
    \midrule
    Top ICD $\uparrow$ & $0.954 \pm 0.1$ & $0.953 \pm 0.1$ & $0.204 \pm 0.4$ & $0.948 \pm 0.2$ \\
    $|\Delta|$ Other ICD $\downarrow$ & $\mathbf{0} \pm \mathbf{0}$ & $21.7 \pm 123.3$ & $372.8 \pm 281.4$ & $9.9 \pm 5.0$ \\
    \bottomrule
    \end{tabular}}
    \captionsetup{name=Table}
    \caption{Ablating the class-specific weights in $\mathbf{A_{f_\text{icd}}}$  does not affect other classes compared to relevant token perturbations, indicating its extremely precise explanation of downstream code predictions.}
    \label{tab:Ablation Study}
\end{table}

\textbf{Global vs. Local Explanations.} Local explanation methods can identify relevant tokens for a model's predictions but fail to isolate the mechanisms behind individual ICD code predictions in extreme multilabel settings, where medically-specific tokens often relate to multiple ICD codes. Our sparse weight matrix $\mathbf{A_{f_\text{icd}}}$ overcomes this limitation by directly mapping encoded dictionary features $\mathbf{f}$ to each ICD code prediction. Through weight ablation, we can pinpoint the specific mechanisms driving each code's prediction. Table \ref{tab:Ablation Study} demonstrates that ablating relevant weights in our global $\mathbf{A_{f_\text{icd}}}$ matrix does not affect the prediction of other ICD codes, unlike token ablation, which impacts multiple codes due to tokens' relationships with various codes.

\begin{figure}[h] 
\centering
\includegraphics[width=0.9\textwidth]{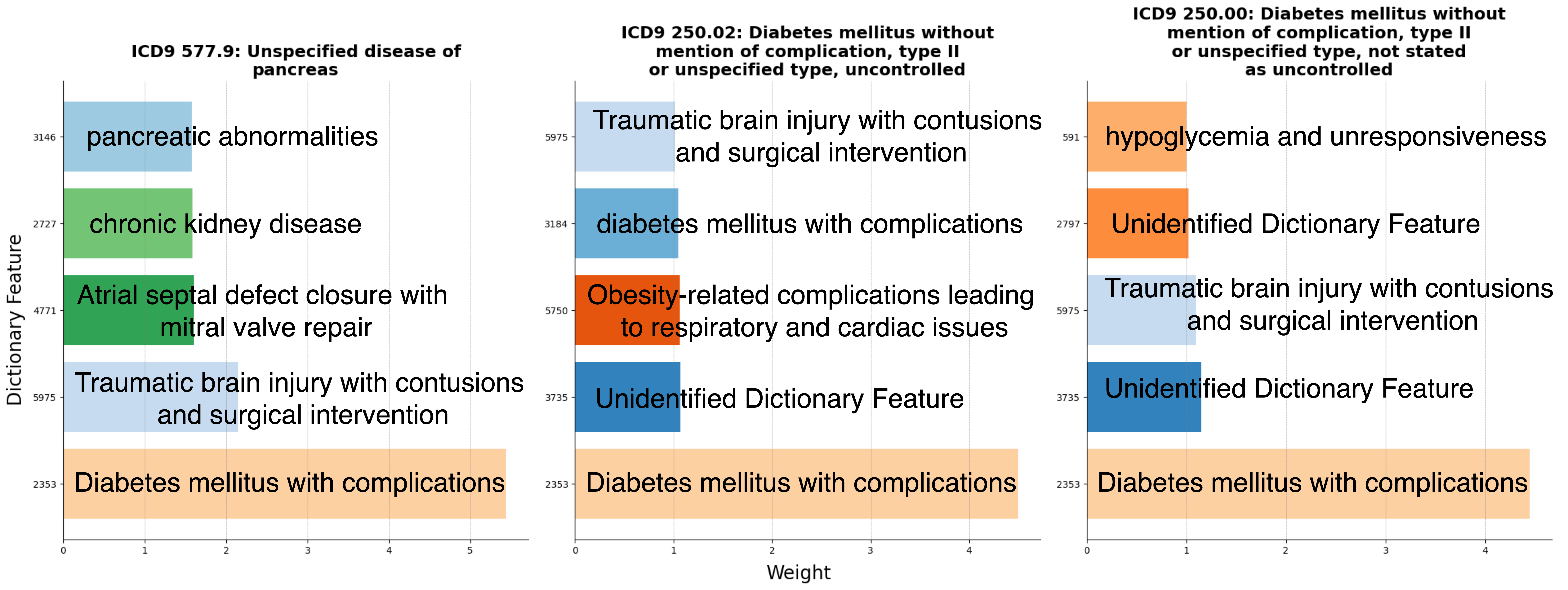}
\caption{Top 5 Dictionary Features for Diabetes-related ICD Codes.}
\label{fig:DiabetesBar}
\end{figure}

\begin{figure}[h] 
\centering
\includegraphics[width=0.9\textwidth]{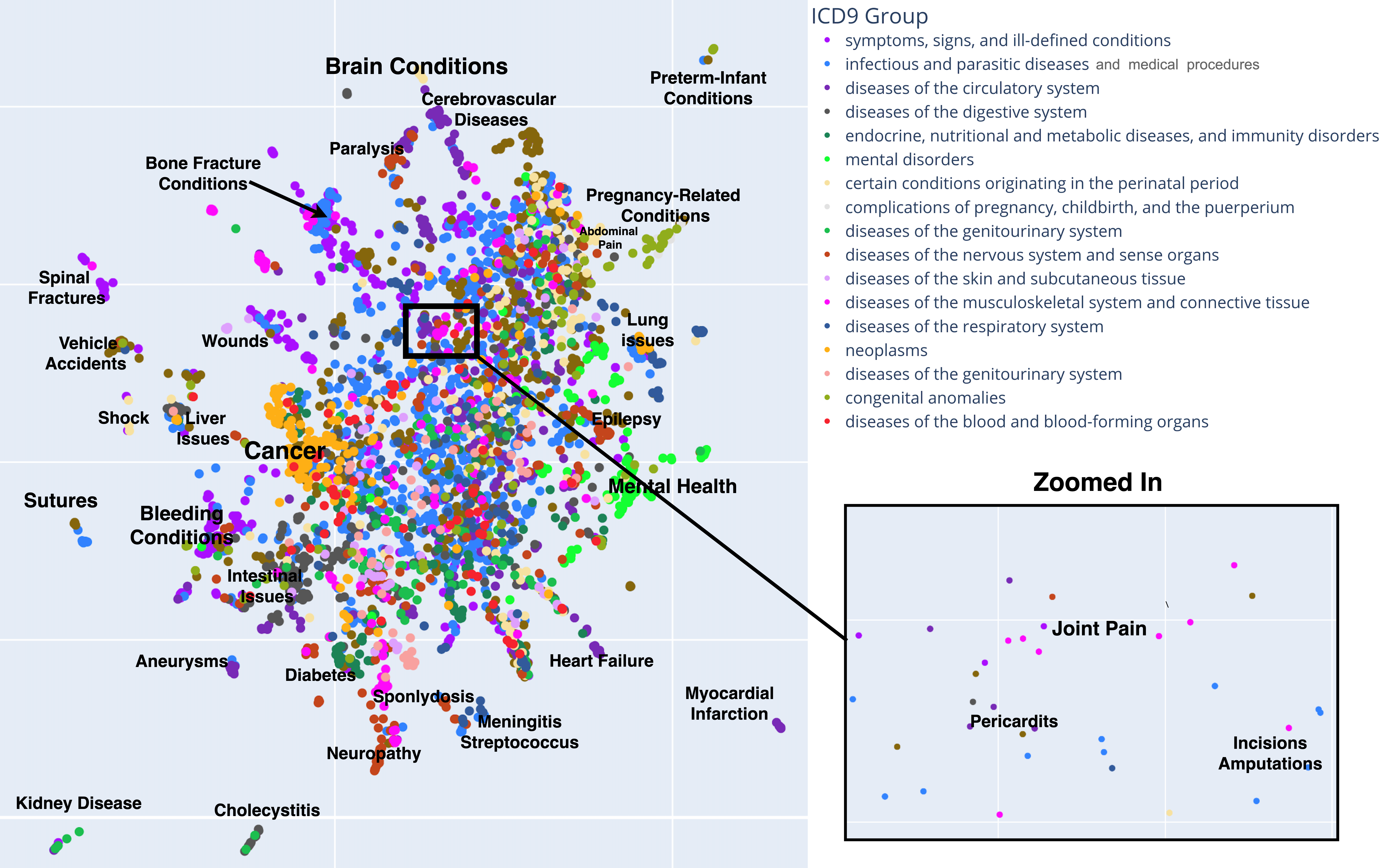}
\caption{UMAP of $\mathbf{A_{f_\text{icd}}}$ with respect to all medical codes. We observe clusters of medical codes with relative distances that are intuitive. For instance, neuropathy is a common condition associated with diabetes, and vehicle accidents are more closely linked to bone and spinal fractures. }
\label{fig:UMAP}
\end{figure}

\textbf{Visualization.} Another crucial utility is we are now able to visualize the overall medical concepts that the model has learned to associate with each of the several thousand ICD codes. For instance, we can summarize that the model has learned that hypoglycemia, obesity, pancreatic abnormalities, and diabetes mellitus is highly predictive of diabetes-related ICD codes in Figure \ref{fig:DiabetesBar}. Furthermore, we can even visualize the model's understanding of different medical conditions in space by plotting the UMAP of $\mathbf{A_{f_\text{icd}}}$ as shown in Figure \ref{fig:UMAP}.

\textbf{Setup.} To assess the interpretability of the global projection matrix $\mathbf{A_{f_\text{icd}}}$, we conduct a predictability experiment. Our medical experts are asked to choose the set of codes that best match a given dictionary feature's LLM summary and its sampled contexts from two sets of codes. They also are asked of their confidence in their choices. Clear and easily understandable dictionary features should represent distinct medical concepts with obvious associated medical codes. The experiment is conducted for 100 randomly sampled dictionary features and repeated using our medical LLM to interpret a larger portion of the matrix. 

\textbf{Baseline.} For a direct comparison, we perform the same experiment using the dense $\mathbf{W_c}$ matrix with the identified features of $\mathbf{Z}$ from Section \ref{results: auto interp}. However, as a quick caveat, due to the larger number of identified abstractions of $\mathbf{f}$ compared to $\mathbf{Z}$, $\mathbf{A_{f_\text{icd}}}$ is inherently more informative. Random guessing in this experiment would yield a 50\% accuracy, as it is effectively a binary classification task. 

\begin{table}[h]
\centering
\resizebox{\textwidth}{!}{
\begin{tabular}{lcccccc}
\toprule
& \makecell{Medical Expert 1 \\Matching Accuracy (100) $\uparrow$} & \makecell{Medical Expert 2 \\Matching Accuracy (100) $\uparrow$} & \makecell{Medical Expert 1 \\Confidence (100) $\uparrow$} & \makecell{Medical Expert 2 \\Confidence (100) $\uparrow$} & \makecell{LLM Matching \\Accuracy (100) $\uparrow$} & \makecell{LLM Matching \\Accuracy (all) $\uparrow$} \\
\midrule
$\mathbf{A_{f_\text{icd}}}$ & \textbf{0.73} \textbf{(+12.3\%)} & \textbf{0.69} \textbf{(+15.0\%)} & $\textbf{3.41} \pm \textbf{0.6}$ & $\textbf{2.87} \pm \textbf{1.0}$ & \textbf{0.65} \textbf{(+6.6\%)} & \textbf{0.59} \textbf{(+9.3\%)} \\
$\mathbf{W_c}$ & 0.65 & 0.60 & $3.20 \pm 0.57$ & $2.62 \pm 0.89$ & 0.61 & 0.54 \\
\bottomrule
\end{tabular}}
\captionsetup{name=Table}
\caption{Human Predictability Experiment with $\mathbf{A_{f_\text{icd}}}$. Our medical experts more confidently and accurately match the corresponding set of ICD codes given a dictionary feature. The relative improvement of $\mathbf{A_{f_\text{icd}}}$ over $\mathbf{W_c}$ is shown in parentheses.}
\label{tab:MedCodeAssociations}
\end{table}
\textbf{Human Evaluations.} From Table \ref{tab:MedCodeAssociations}, we observe that $\mathbf{A_{f_\text{icd}}}$ is more interpretable than $\mathbf{W_c}$ due to a larger portion of codes being matched when using features from $\mathbf{f}$ compared to $\mathbf{Z}$, but with some interesting caveats that make human predictability challenging. Shown in Appendix Figure \ref{fig: f_Human_predict}, many of the incorrect matchings by humans were due to incorrectly learned associations between the identified abstract dictionary features and its corresponding top 5 medical codes despite the $\mathbf{f}$ generally being interpretable themselves, suggesting that $\mathbf{A_{f_\text{icd}}}$ may prove to be useful in identifying incorrectly learned feature predictions.

\vspace{-0.2cm} 
\subsection{Performance} \label{sec: perf}
\textbf{Baselines.} The trade-off between interpretability and performance due to reduced expressive power is a critical concern when integrating sparse layers and more interpretable modules \cite{mansour2022accuracyinterpretability_tradeoff, Rudin2019_InterpretableModelsTradeoff}. However, our findings suggest that this trade-off may not persist. Table \ref{tab:methods} compares the performance of various automated ICD coding methods such as convolutional neural networks, recurrent neural networks, and PLMs for MIMIC-III clean \cite{AutomatedMedicalCoding}, including our own reproductions of PLM-ICD (*) as a baseline.

\begin{table}[h]
\centering
\resizebox{\textwidth}{!}{%
\begin{tabular}{lcccccc|cc}
\toprule
 & CNN & Bi-GRU & CAML \cite{mullenbach2018explainable_CAML} & MultiResCNN \cite{Li2020-iz-MRCNN} & LAAT \cite{Vu_2020_LAAT} & PLM-ICD \cite{huang2022plmicd} & PLM-ICD* & \method* (ours) \\
\midrule
Micro F1 $\uparrow$ & $48.0 \pm 0.3$ & $49.7 \pm 0.4$ & $55.4 \pm 0.1$ & $56.4 \pm 0.2$ & $57.8 \pm 0.2$ & $59.6 \pm 0.2$ & $54.6 \pm 0.1$ & $54.9 \pm 0.2$ \\
Macro F1 $\uparrow$ & $9.9 \pm 0.4$ & $12.2 \pm 0.2$ & $20.4 \pm 0.3$ & $22.9 \pm 0.6$ & $22.6 \pm 0.6$ & $26.6 \pm 0.8$ & $26.5 \pm 0.3$ & $ 27.2 \pm 0.4$ \\
Micro AUC-ROC $\uparrow$ & $97.1 \pm 0.0$ & $97.8 \pm 0.1$  & $98.2 \pm 0.0$ & $98.5 \pm 0.0$ & $98.6 \pm 0.1$ & $ 98.9 \pm 0.0$ & $97.7 \pm 0.0$ & $97.6  \pm 0.0 $\\
Macro AUC-ROC $\uparrow$ & $88.1 \pm 0.2$ & $91.1 \pm 0.2$  & $91.4 \pm 0.2$ & $93.1 \pm 0.3$ & $94.0 \pm 0.3$ & $95.9 \pm 0.1$ & $92.5 \pm 0.0 $ & $91.7  \pm 0.0 $\\
\bottomrule
\end{tabular}
}
\caption{Performance comparison of automated ICD coding methods on MIMIC-III clean dataset \cite{AutomatedMedicalCoding}. * indicates our training or reproduction. The average scores are reported along with their standard deviations.}
\label{tab:methods}
\end{table}

\textbf{Performance.} Our model, \method, achieves the highest average Macro F1 score and slightly lower Micro F1 score compared to the relevant baselines, indicating better performance on rarer codes but worse performance on edge-cases of common ICD codes. Notably, our reproduction of the previous state-of-the-art baseline, PLM-ICD, yields lower performance than reported, possibly due to memory restrictions limiting batch sizes. As batch size directly affects the optimal performance of ICD coding models \cite{AutomatedMedicalCoding}, \method may still attain better performance with larger batch sizes.

%% file: sections/5conclusion.tex
\vspace{-0.2cm} 
\section{Discussion and Conclusion} \label{sec: Discussion}
\vspace{-0.2cm} 
\begin{wraptable}{r}{0.45\textwidth}
    \centering
    \begin{tabular}{lcc}
    \hline
    & Pre-Edit & Post Edit \\
    \midrule
    False Positives & 164 & 158 \\
    False Negatives & 66 &  68 \\
    \bottomrule
    \end{tabular}
    \captionsetup{name=Table}
    \caption{Results of causal edits of $A_{f_{icd}}$ for ICD 99.20 "Injection of Platelet Inhibitors".}
    \label{tab:Debugging}
\end{wraptable}

\textbf{Debugging.} Global interpretability allows for quick identification of incorrectly learned mappings between medical concepts and codes by inspecting the sparse mappings in $A_{f_{icd}}$. We have observed numerous improperly learned associations, as visualized in Figures \ref{fig: f_heatmap_chest_tube}, \ref{fig: f_heatmap_sepsis}, and \ref{fig: f_heatmap_diabetes} in the Appendix. For example, the code for "Athlete's Foot" is incorrectly associated with Ovarian cancer in Figure \ref{fig: f_heatmap_sepsis}, potentially leading to false positives. In a case study, we attempted to remedy the commonly false positive code "Injection of Platelet Inhibitors" by visualizing its top 20 most common abstract dictionary features in Figure \ref{fig: f_platelet_bar}, revealing incorrect associations like dictionary feature 4443 "Trisomy Disorders" (see Section \ref{Appendix: Debugging}). Ablating the relevant weights in $A_{f_{icd}}$ decreased false positives related to the incorrect dictionary features but slightly increased false negatives (Table \ref{tab:Debugging}), suggesting the need to change other weights to better predict true cases, and highlighting the complexities of debugging. An interpretable automated procedure for debugging incorrectly classified ICD codes is an important direction, as over 40\% of ICD codes are never predicted correctly~\cite{AutomatedMedicalCoding}.

\textbf{Improving LLM Annotations.}  Our current automatic interpretability pipeline with LLMs only uses zero-shot prompting. However, numerous works have improved LLM-assisted annotations \cite{goel2023llms_accelerate_medical_information_extraction} and generation faithfulness such as retrieval augmented generation \cite{lewis2021retrievalaugmented}. Exploring these ideas could bridge the gap between current state-of-the-art domain-specific LLMs and medical experts for dictionary feature annotation tasks.

\textbf{Unidentifiable Dictionary Features.}  Some highly relevant dictionary features may not be identifiable by humans or LLMs, as shown in the Appendix (Figure \ref{fig: f_heatmap_diabetes}). We investigated a few and showcase their sampled contexts in the Appendix (Table \ref{tab: Unidentifiable Dictionary Features}). Unidentified features often result from a lack of highly activated contexts within our text corpus or a lack of an explicit coherent medical theme, suggesting the need for larger sampled corpuses in our dictionary construction and potential limitations in our dictionary learning formulation. Other sparse formulations should be explored for optimal interpretable design \cite{rajamanoharan2024improvingSAE}.

Ultimately, our proposed dictionary label attention (\method) module takes a step towards addressing the need for interpretability in high-dimensional multilabel prediction tasks, particularly in medical coding. By disentangling dense embeddings into a sparse space and leveraging LLMs for automated dictionary feature identification, \method aims to uncover globally learned medical concepts, provide comprehensive explanations, and facilitate the development of debuggable models. While further research is needed to validate its effectiveness, \method represents a promising direction in developing more interpretable and transparent models for complex, high-stakes applications, contributing to developing trustworthy AI systems in healthcare and beyond.


%% file: sections/6appendix.tex
\subsection{Training Details} \label{Appendix : Training Details}
For a fair comparison and to reproduce the PLM-ICD model from \cite{AutomatedMedicalCoding}, we followed the hyperparameters swept in their work as closely as possible. Specifically, both models employed the same pre-trained medical RoBERTa encoder architecture. One slight difference in training is that we leverage the PLM already pre-trained on the ICD coding task whereas PLM-ICD is trained from a pre-trained medical RoBERTa model on other medical data. Additionally, we utilized their code to perform the same data splits as reported in their MIMIC-III reproduction, obtained from their \href{https://github.com/JoakimEdin/medical-coding-reproducibility}{repository} \cite{AutomatedMedicalCoding}. Due to physionet's policies, we are unable to share their training data directly. However, due to GPU memory constraints, we were unable to use the same batch size, which may have hindered our ability to fully reproduce their PLM-ICD \cite{huang2022plmicd} performance. Table \ref{tab:hyperparameters} showcases the hyperparameters used for our methods. It's worth noting that an initial parameter sweep revealed that the performance benefits of PLM label attention models largely stemmed from their batch size. Consequently, we reran our training four times across four randomly selected seeds. We use the optimizer AdamW with its default settings. We also use 1e-6 for our $\lambda_{\text{saenc}}$ hyperparameter.

\begin{table}[h]
\centering
\caption{Hyperparameters used in training DILA and PLM-ICD models}
\label{tab:hyperparameters}
\resizebox{\textwidth}{!}{
\begin{tabular}{lcccccccc}
\toprule
& Learning rate & $\lambda_{\text{l1}}$ & $\lambda_{\text{l2}}$ & Batch size & LR scheduler & Epochs & Dropout & Decision Boundary Threshold  \\
\midrule
DILA *& 5e-5 & 0.0001 & 0.00001 & 8 & Linear Warmup & 20 & 0.2 & 0.3 \\
PLM-ICD* & 5e-5 & N/A & N/A & 8 & Linear Warmup & 20 & 0.2 & 0.3 \\
PLM-ICD \cite{AutomatedMedicalCoding} & 5e-5 & N/A & N/A & 16 & Linear Warmup & 20 & 0.2 & 0.3 \\
\bottomrule
\end{tabular}
}
\end{table}

\textbf{Compute Resources.} We leverage a compute cluster using A6000 48GB GPUs and note that our maximum batch size was of size 8 due to some clinical notes containing over 4,000 tokens. Training takes approximately a day, but we regret not measuring explicit training times. We recommend having at least 48GB of CPU memory, as depending on the analysis, we store a lot in memory.

\subsection{Dictionary Contexts Construction} \label{Appendix: Dictionary Construction}
The construction of a human-interpretable dictionary, containing relevant tokens for each $f_i$, involves two steps. First, we sample a large number of tokens and decompose their embeddings using the sparse autoencoder defined in Section \ref{sec: DL}. For each $i$ in $\mathbf{f}$, we retrieve the sparsely activated nonzero $f_i$. Next, we sort all tokens by their $f_i$ magnitudes for each $f_i$ to obtain their contexts. To conserve memory, we save the top 10 connected tokens but only use the top 4 connected tokens (i.e., multiple activated tokens within a chunk) for our evaluations. It's worth noting that some dictionary features have very few activating contexts. Algorithm \ref{alg:build_dict} summarizes this procedure. For our dictionary construction, we sample over 8,000 clinical notes from the test set.
\begin{algorithm}[h!]
\caption{Build Dictionary}
\label{alg:build_dict}
\KwIn{Sparse Autoencoder $A$, feature $f_i$, tokens $x$}
\KwOut{Dictionary $\mathbf{f}$ mapping $f_i$ to tokens $x$ and classes $y$}

$F \gets \text{dict}$;
\For{each token $x$}{
$f \gets A.encode(x)$;
\For{$f_i$ in $\mathbf{f}$}{
\If{$f_i > F[i].f_i$}{
$F[i].tokens \gets x$;
}
$\delta_i \gets \text{ablation}(f_i)$;
\If{$\delta_i > F[i].\delta$}{
$F[i].classes \gets \text{drops}(\delta_i)$;
}
}
}
\Return $\mathbf{f}$;
\end{algorithm}

\begin{figure}[H] 
\centering
\includegraphics[width=0.8\textwidth]{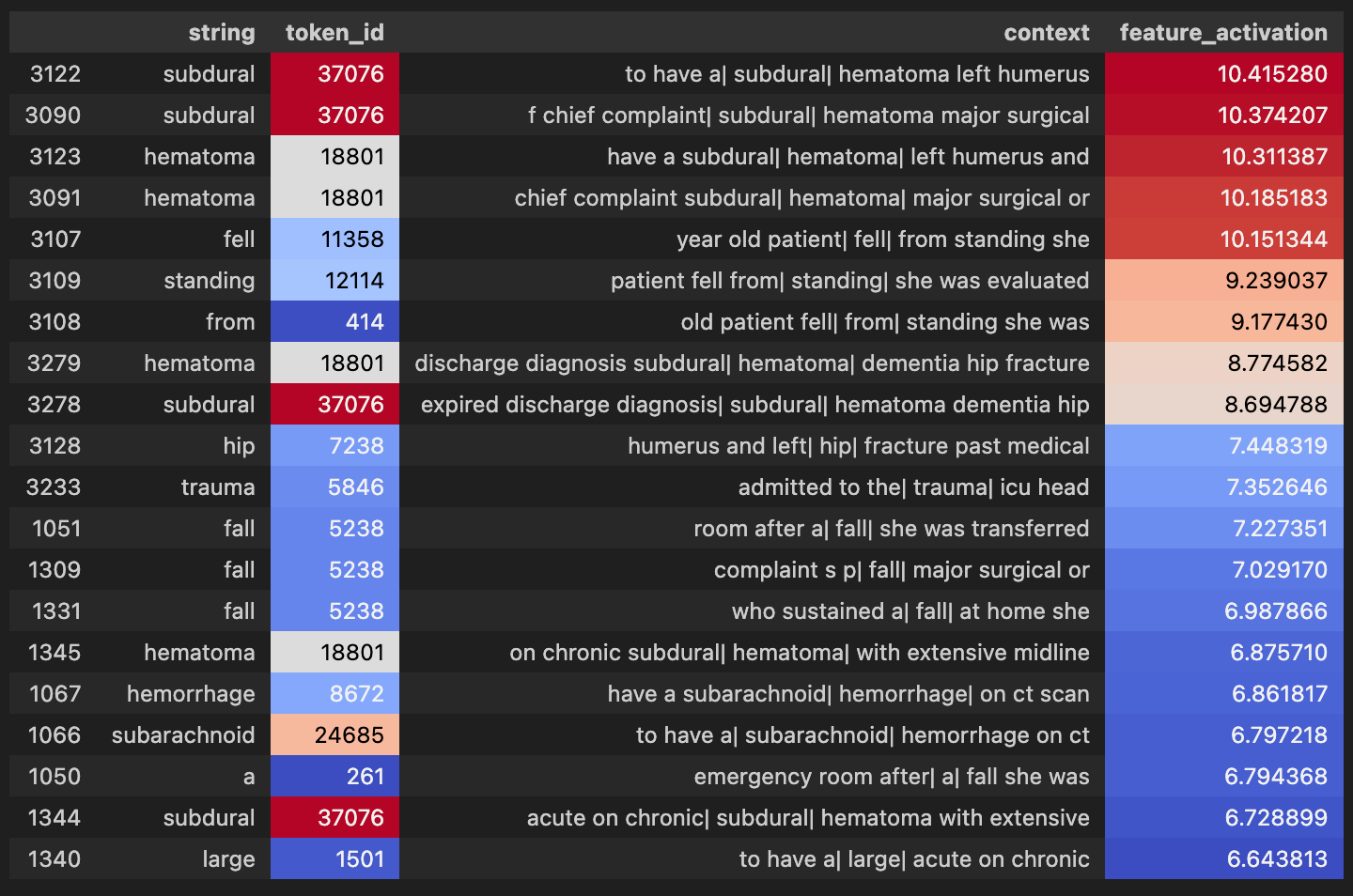}
\caption{Example of token sorting to acquire the necessary token contexts for $f_i$ for dictionary feature 1,871, relating to subdural hemotomas. The far left column indicates the position of the token in the text corpus.}
\label{fig:sorting}
\end{figure}

We show a resulting output of a pandas dataframe in Figure \ref{fig:sorting}.

\subsection{Dictionary Feature Identification}
\textbf{Extended Rationale for Dictionary Feature Identification Approach with LLMs:} Initially, we planned to have the LLM simply answer whether or not there was an explicit medical theme within each set of contexts with a yes or no response. However, we observed that, regardless of specificity, the LLMs would always respond affirmatively, stating that the contexts were medically relevant. Consequently, we pursued a different approach originally leveraged in human evaluation experiments by \cite{subramanian2017spine}. We noticed that this approach substantially improved the determination of whether a set of dictionary contexts was interpretable, as the LLM was implicitly forced to evaluate whether a common concept existed among the dictionary contexts. To illustrate the dictionary feature identification process, we showcase a couple of LLM prompts and their respective context tokens in the figures below.

\textbf{Other Minor Identification Experiment Details:} We ran our dictionary feature identification process across 6,088 dictionary features. Initially, we eliminated any dictionary feature with fewer than 4 contexts, deeming them unidentifiable due to the lack of contexts. Subsequently, we conducted our simple identification experiment. It's worth noting that, in principle, larger text corpuses could potentially be beneficial for identifying more dictionary features. However, the majority of dictionary features (5,847 out of 6,088) contained at least 4 contexts.

\subsubsection{LLM Identification Prompt} \label{appendix: LLM identfication}
\begin{figure}[H] 
\centering
\includegraphics[width=0.8\textwidth]{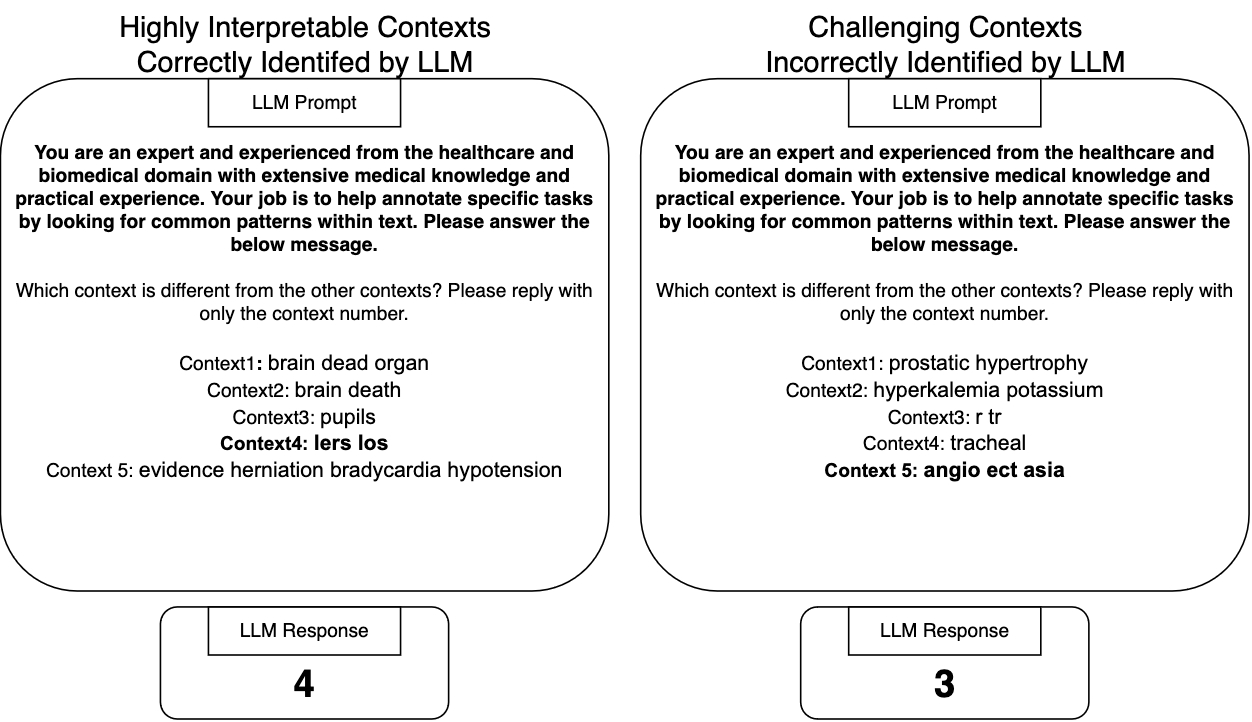}
\caption{Dictionary Feature $f_i$ Identification Tasks for the LLM. The correct random context is indicated in bold. LLMs struggle to identify the common medical theme in challenging contexts that lack explicit connections (right). Clinical notes often contain abbreviations and shorthand without clear references, making interpretation difficult even for experienced physicians. For example, the abbreviation "rtr" was challenging for our clinical physician to recall (right). }
\label{fig: L1_LLM_ID}
\end{figure}

\begin{figure}[H] 
\centering
\includegraphics[width=0.8\textwidth]{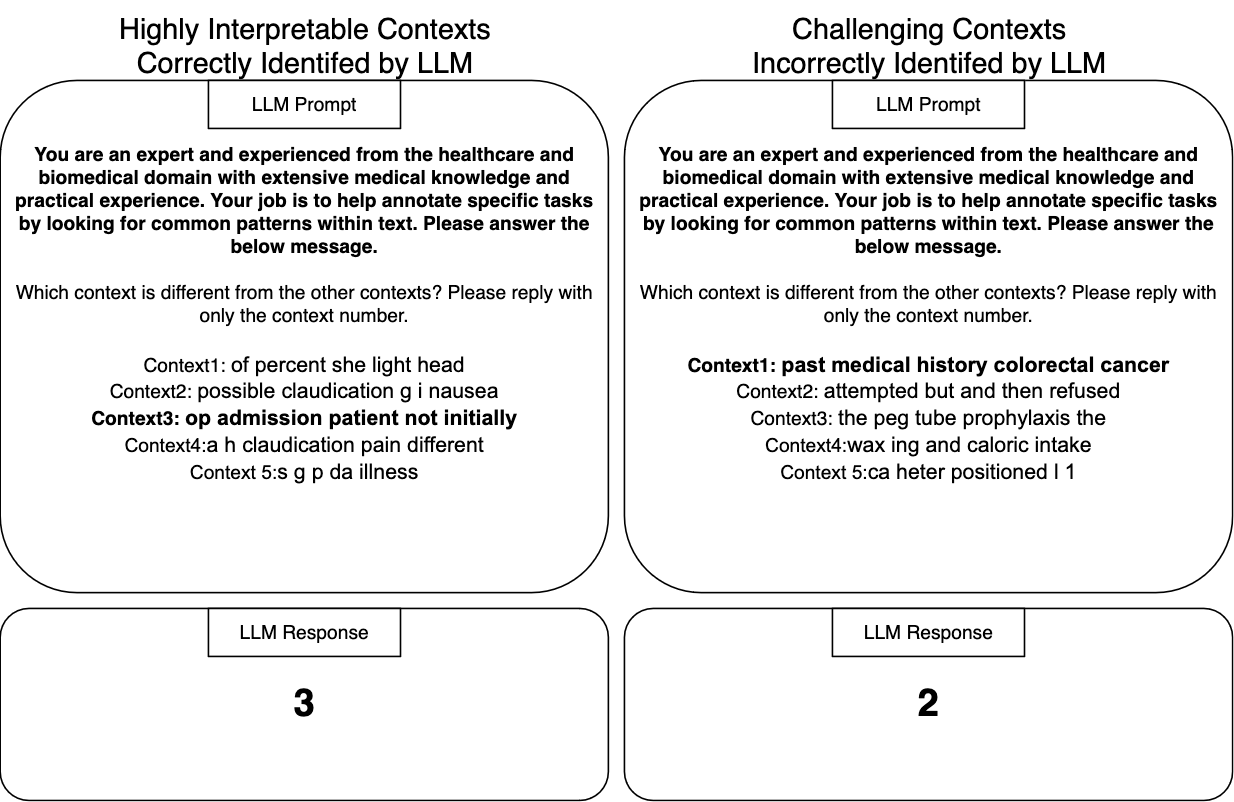}
\caption{Dense $\mathbf{Z_i}$ Identification Tasks for the Language Model (LLM). The correct random context is indicated in bold. Some dense layer neurons activate for contexts with a common theme, such as claudication and nausea (left). However, many neurons activate for seemingly unrelated contexts (right).}
\label{fig: Z_LLM_ID}
\end{figure}

\subsubsection{Human Identification Examples}
\begin{figure}[H] 
\centering
\includegraphics[width=0.8\textwidth]{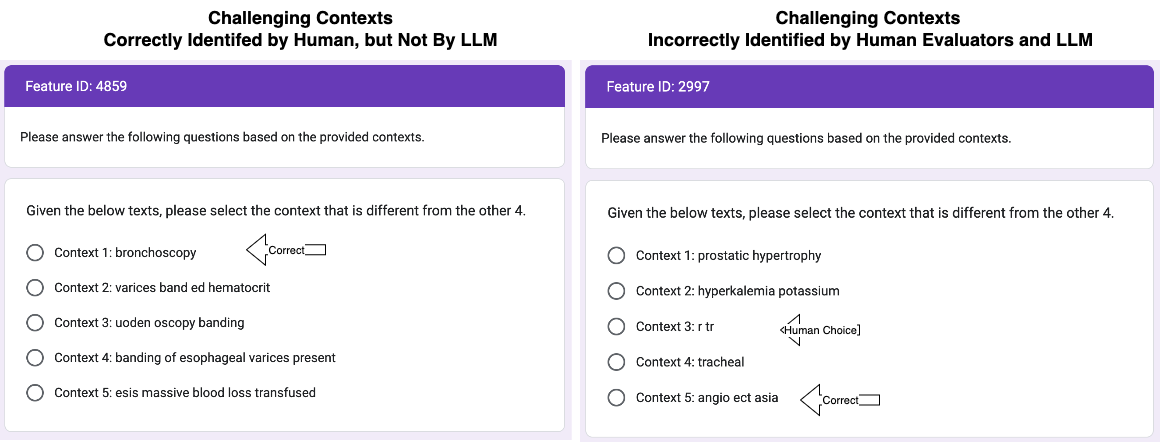}
\caption{Dictionary Feature $f_i$ Identification Tasks for Humans. The left panel shows a case where the human correctly identified a dictionary feature, but the Language Model (LLM) failed to do so. The LLM incorrectly selected "massive blood loss..." as the random context, despite its relationship to banding. The right panel presents dictionary contexts where both the LLM and humans failed to identify the randomly sampled context, which is understandable given the seemingly unrelated nature of the contexts. }
\label{fig: f_human_ID}
\end{figure}

\begin{figure}[H] 
\centering
\includegraphics[width=0.8\textwidth]{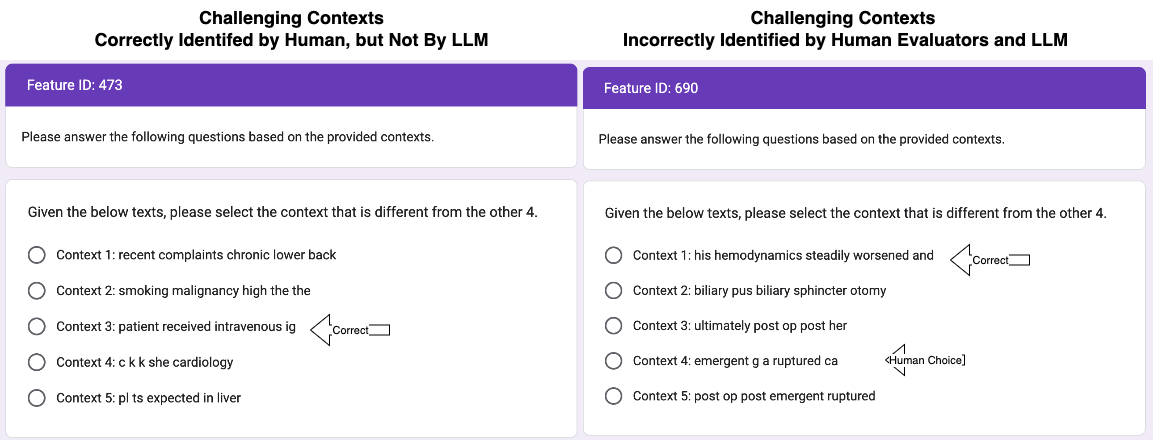}
\caption{Dense $\mathbf{Z_i}$ Identification Tasks for Humans. The left panel shows an example where humans identified a neuron with hidden underlying relationships, such as the connection between smoking and cardiology, where patients often experience various levels of pain due to complications. Language Models (LLMs) struggle to identify such relationships. The right panel demonstrates a neuron with a diverse set of contexts, making it challenging for both humans and LLMs to identify a common theme.}
\label{fig: Z_human_ID}
\end{figure}

\begin{figure}[H] 
\centering
\includegraphics[width=0.4\textwidth]{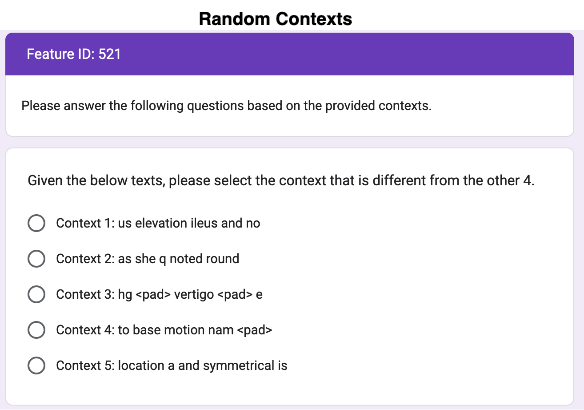}
\caption{Example of Purely Random Contexts. The image showcases a prompt with truly random tokens, including <pad> tokens that were not filtered out. A stark reduction in <pad> tokens activating dense $\mathbf{Z}$ neurons is observed, and <pad> tokens are essentially never present in the dictionary contexts, suggesting their irrelevance to model predictions.}
\label{fig: Z_human_ID}
\end{figure}

\subsection{LLM Summarization}
\begin{figure}[H] 
\centering
\includegraphics[width=1.15\textwidth]{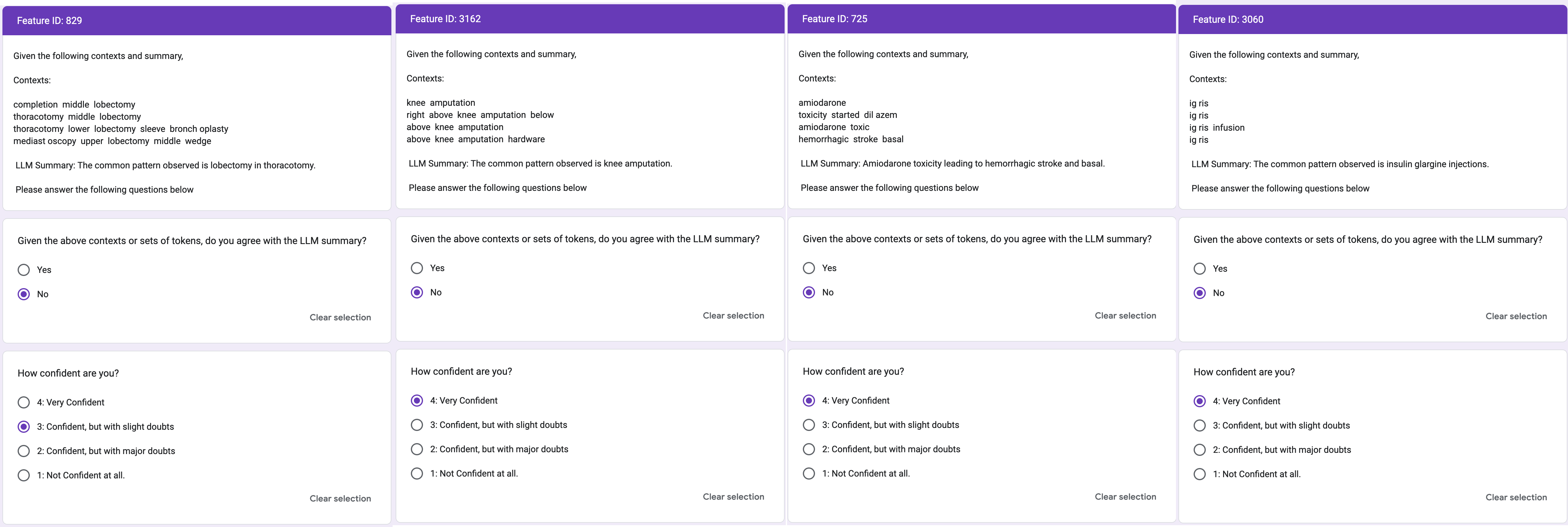}
\caption{Medical Expert Rejected LLM Summaries of Dictionary Features $\mathbf{f}$. The summaries rejected by the medical experts were generally due to a lack of specificity, such as "knee amputation" not being "above knee amputation" (middle left). Two other summaries (left and middle right) were rejected because the LLM summary was too specific, with unlikely associations such as Amiodarone and strokes or the assumption that thoracotomy was present in all contexts. The far right rejection demonstrates a case of direct hallucination, where the LLM assumed the contexts were related to insulin despite no information hinting at that relationship, even though the contexts shared the same acronym. }
\label{fig: f_human_summary}
\end{figure}

We showcase the LLM prompt used, and the human evaluation results below. We showcase the summaries rejected by our medical experts in Figure \ref{fig: f_human_summary}.
\subsubsection{LLM Prompt}
\begin{figure}[H] 
\centering
\includegraphics[width=0.4\textwidth]{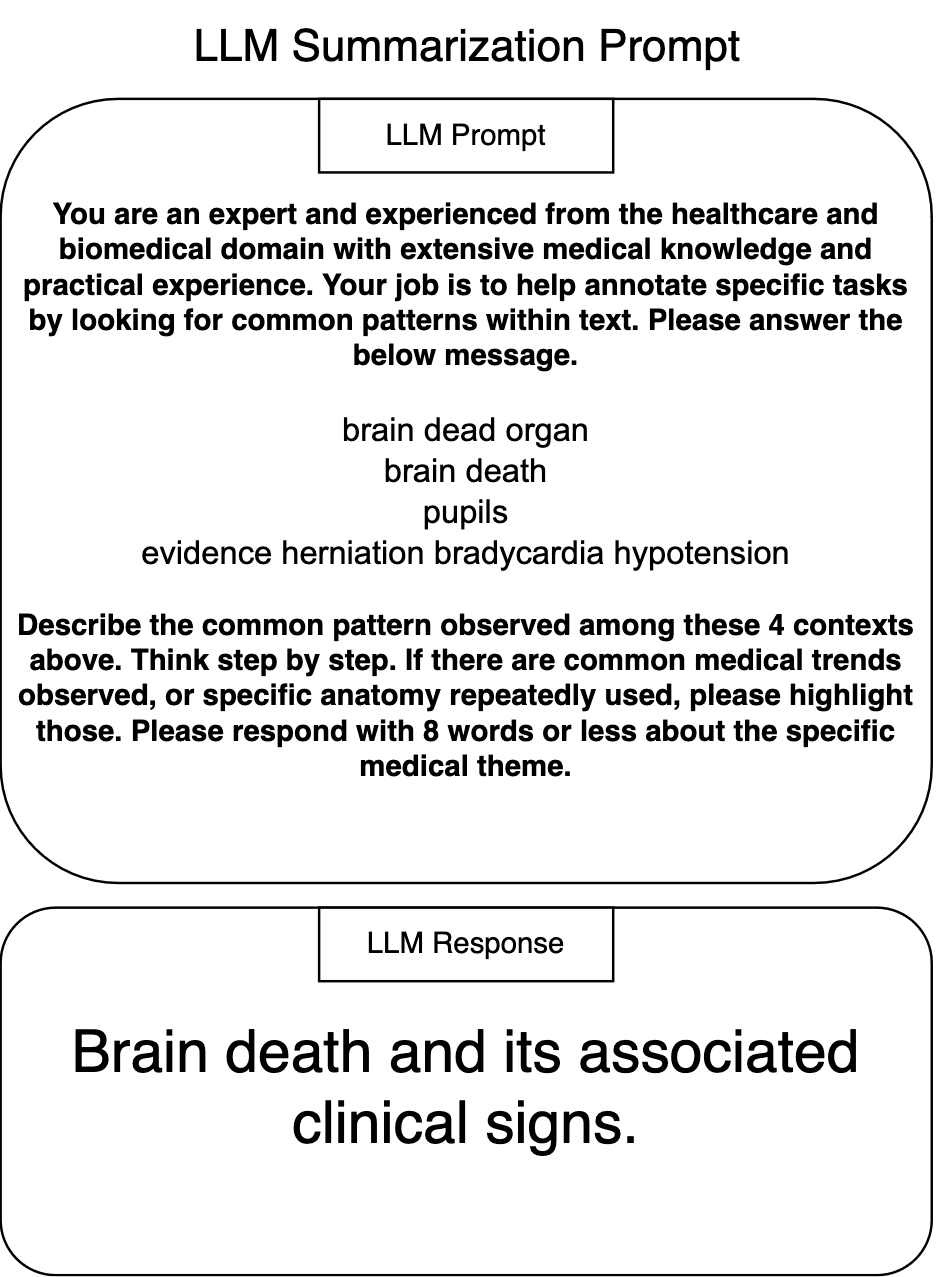}
\caption{LLM Summarization Prompt for Dictionary Feature Contexts. The image shows the prompt used to summarize the highly connected dictionary contexts identified in the pipeline. The model is limited to 8 words to facilitate processing and extracting the summaries, as allowing an unlimited word count resulted in the inclusion of filler words in the LLM-generated summaries.}
\label{fig: f_LLM_summary}
\end{figure}

\subsubsection{Medical Expert Rejections of LLM Summary} \label{Appendix: Human Summary Eval}
We showcase only the LLM-generated summaries rejected by our medical expert evaluators, as they are more informative than the accepted summaries. Although the dictionary contexts are highly interpretable and consistent with a specific medical theme, the experts disagreed with certain aspects of the specific LLM summaries. For instance, the term "knee amputation" was deemed insufficiently specific, as "above knee amputation" is the more precise medical condition, differing from "below knee amputations." Other disagreements were related to the LLM's inferences of abbreviations, such as "ig ris," where there was no evidence of "insulin glargine injections" within the context. In such scenarios, dictionary contexts may not have a conclusive summary, requiring further investigation into the clinical notes as well as potentially the need to include some amount of context. Note that, as part of the LLM auto-interpretability evaluation study done in \ref{tab:identification test}, 59 dictionary feature summaries and 34 Dense Z summaries were examined, as shown in Table \ref{tab:summary agreement}.

On the other hand, from the dense $Z$ contexts that have seemingly passed the LLM identification test, their summaries were rejected due to the lack of coherence within each context, implying that many of the identified $Z$ features may have been the result of chance by the LLM pipeline. 

\begin{figure}[H] 
\centering
\includegraphics[width=1.15\textwidth]{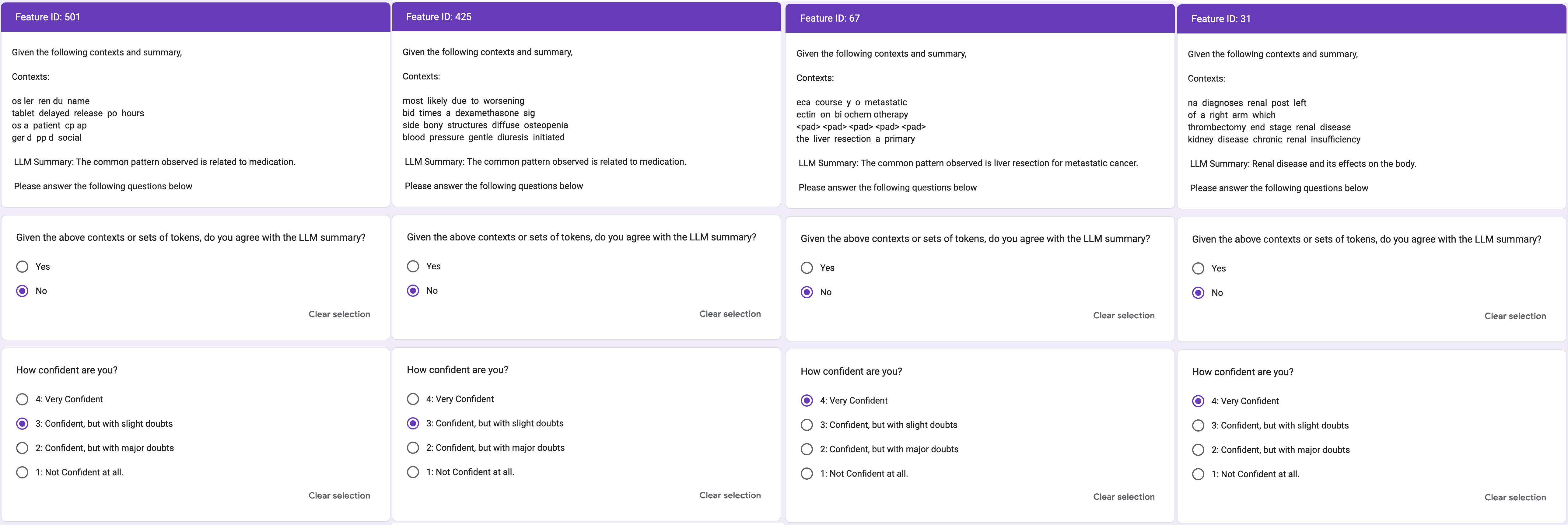}
\caption{Medical Expert Rejected LLM Summaries of Contexts from Dense $Z$. Many of the dense $Z$ context summaries were rejected due to their over-generality. For instance, the LLM would frequently generate summaries like "the common pattern observed is related to medication," which is true to some extent as medical text is generally related to medication. However, these LLM summaries (far left, middle left) were deemed uninformative by the medical experts. Other rejections (middle right) were due to the lack of coherence within dense $Z$ contexts, such as the presence of <pad> tokens and liver-related contexts. Surprisingly, the LLM summary still managed to capture some relevant parts of the context, such as metastatic cancer and liver resection. Finally, the far right example shows a surprisingly consistent medical theme, with renal disease present in three of the four contexts. However, the annotators felt that the summary "effects on the body" was too broad and not specific enough. }
\label{fig: Z_human_summary}
\end{figure}

\subsection{Sparse Projection Matrix} \label{Appendix: Sparse Projection Matrix}
To better understand the structure of our sparse matrix, we visualize the first 100 dictionary features and 100 medical codes in our learned $\mathbf{A_{f_\text{icd}}}$ and dense $W_c$ projection matrix in PLM ICD. We note that our $\mathbf{A_{f_\text{icd}}}$ is substantially sparser.

\begin{figure}[H] 
\centering
\includegraphics[width=0.8\textwidth]{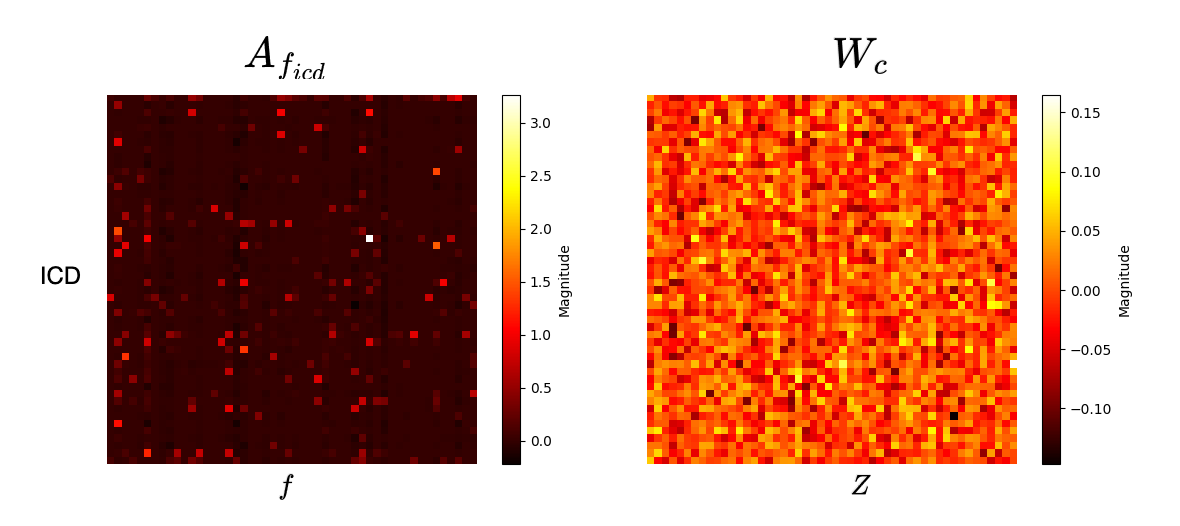}
\caption{Sparsity comparison between ICD projection matrices $W_c$ and $A_{f_{icd}}.$ for first 100 dictionary features and medical codes. Visually speaking, it is easy to identify the strong relationships between dictionary features and ICD codes whereas in the dense projection matrix, its weights look almost uninterpretable.}
\label{fig:dla_projection}
\end{figure}

We leverage this high sparsity to construct our human interpretability and ablation experiments below, as each abstract dictionary feature is observed to be related to specific medical codes. 
\subsubsection{Global vs. Local Interpretability Ablation Experiments}
\textbf{Definition of Highly Relevant Tokens:} Since most tokens have very low label attention attribution scores for each class, we define the most relevant tokens as those with attention attribution scores greater than the 95th quantile for a specific class in the label attention matrix.

\textbf{Weight Ablation Details:} Each clinical note can be decomposed into a sparse set of dictionary features. Consequently, we can identify a set of weights corresponding to each nonzero dictionary feature and medical code to be ablated for each clinical note. Since most irrelevant weights are already close to zero, and the relevant weights are positive, ablating them (i.e., setting them to zero) should provide a reasonably close approximation to the optimal explanation. In practice, ablating all the weights for a specific class in our $\mathbf{A_{f_\text{icd}}}$ matrix does not affect any other class, indicating the potential for precise explanations of a single medical code or class.

\textbf{Recognition of Faithfulness Metrics in Conventional Multiclass Classification:} We acknowledge the existence of various other interpretability faithfulness metrics for local attribution methods \cite{chan-etal-2022-comparative_faithfulness}, such as comprehensiveness \cite{deyoung2020eraser_comprehensiveness} and monotonicity \cite{arya2019explanation_mono}. However, due to the large number of tokens (some clinical notes have upwards of almost 6,000 tokens), performing a quantiling removal or token addition scheme is extremely computationally expensive, especially if one were to consider doing so for each code in every multilabel example. As such, we simplified our downstream effects or faithfulness experiment to simply measuring the drop in performance of the most likely ICD code for each clinical note and the absolute change in softmax probabilities of other ICD code predictions.

\subsection{Predictability Experiments}
In the human predictaiblity experiment, we ask both LLMs and our medical experts to select the best corresponding medical codes given a set of dictionary feature contexts and its LLM summary. We showcase examples of our predictability experiments below. 

\subsubsection{LLM Prompt}
\begin{figure}[H] 
\centering
\includegraphics[width=0.4\textwidth]{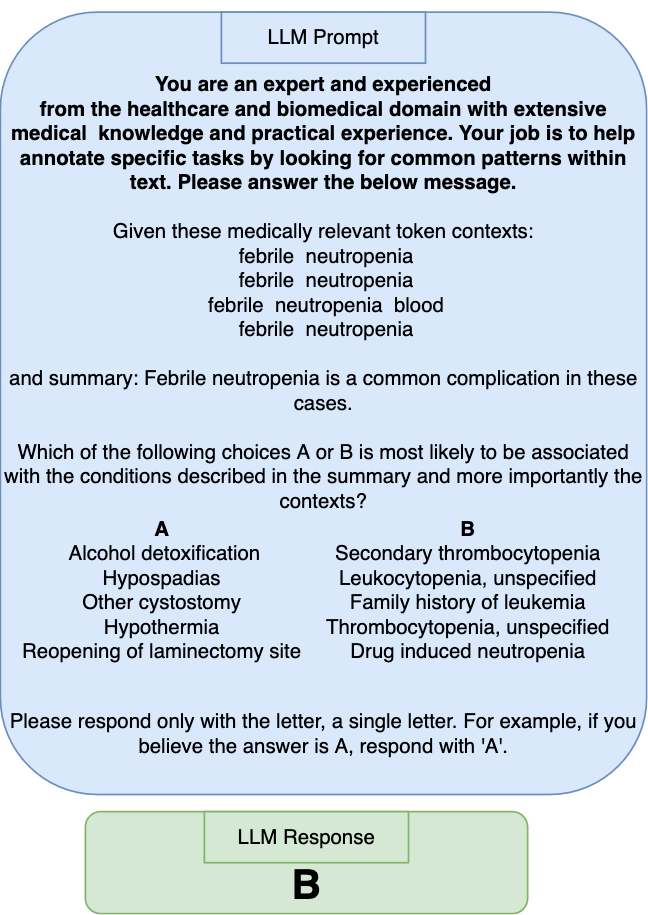}
\caption{LLM Prompt for Predictability Experiment with $\mathbf{f}$. The LLM is presented with two sets of medical code descriptions related to the contexts and summary it generated, labeled as "A" and "B". The LLM is then prompted to choose the set of medical codes that best describes or relates to the conditions shown in the given contexts and summary.}
\label{fig: f_LLM_predict}
\end{figure}

\subsubsection{Human Evaluation} \label{Appendix : Human Predictability Experiment}
Due to being more informative than the positive correct cases (as those dictionary features and codes are more obviously selected for), we showcase the results where the person was unable to select the set of medical codes that the model had highlighted for its downstream predictions. From qualitative inspection with our medical experts, we notice that the medical expert was unable to predict the correct set of medical codes were the direct result of both sets of medical codes being related to the dictionary feature, or the top ICD codes associated with each dictionary feature in our $\mathbf{A_{f_\text{icd}}}$ were incorrectly mapped. For instance, the suture of lacerations of different organs are not directly related to gallstones and double tail stents that prevent biliary obstructions. On the other hand, smoking complications such as Tobacco use are often more likely to cause stomach ulcers than malignant lymphomas. While both can have that as a potential complication, in practice, one is more common over the other, showcasing the wrong weighing of our learned $\mathbf{A_{f_\text{icd}}}$ matrix between different medical conditions and medical codes. Other predictability experiments failed due to the lack of extra information on abbreviations such as "LEVT", which may be a challenge to understand as many of these unstructured clinical notes do not have a specific identification process for these abbreviations. That said, we note that the predictability of these codes is better than random, suggesting that the global dictionary matrix can be leveraged to better understand what our model understands for each set of medical code predictions.
\begin{figure}[H] 
\centering
\includegraphics[width=1.15\textwidth]{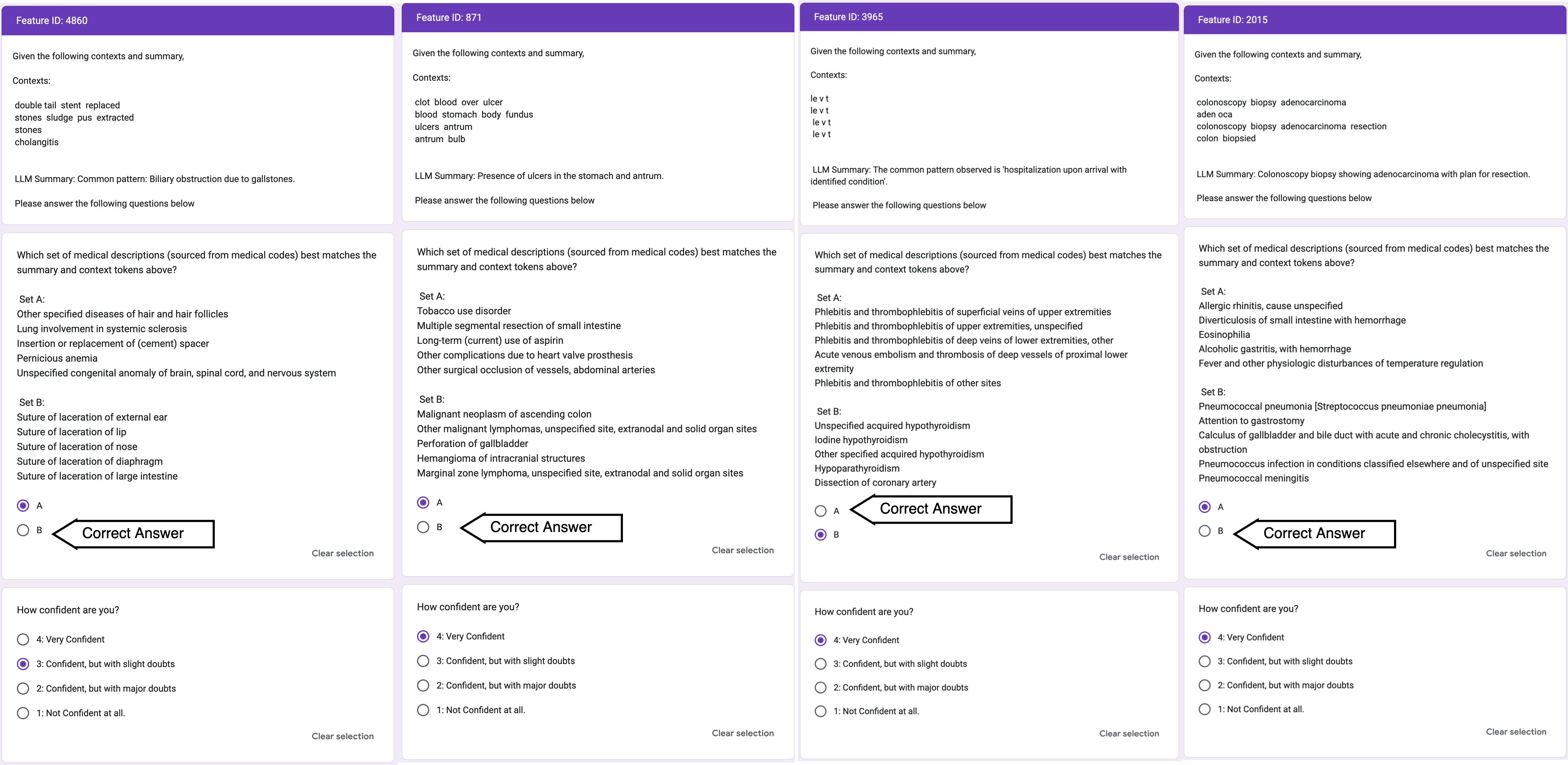}
\caption{Human Predictability Experiment: Cases with incorrect predictions using dictionary features $\mathbf{f}$. Despite many of the contexts being highly informative, the annotator was often unable to select the set of medical codes corresponding to the top 5 medical codes defined by the $\mathbf{A_{f_\text{icd}}}$ matrix. For example, the left contexts are related to stones, which should have no connection to suture of lacerations, but Set B is the corresponding set of codes observed. In the middle left set of contexts, ulcers are a common complication among tobacco users and smokers, but the model has possibly learned, by association, that they are more indicative of cancer. The "LEVT" contexts and the essentially hallucinated summary were virtually unidentifiable despite being extremely consistent in theme. Finally, both sets of medical codes seem to have very little relation to adenocarcinoma (a form of cancer) in the last set of contexts. }
\label{fig: f_Human_predict}
\end{figure}

\subsubsection{Misaligned Associations Between Learned Medical Concepts and Medical Codes} \label{Appendix: misalign}
\begin{figure}[H] 
\centering
\includegraphics[width=1.0\textwidth]{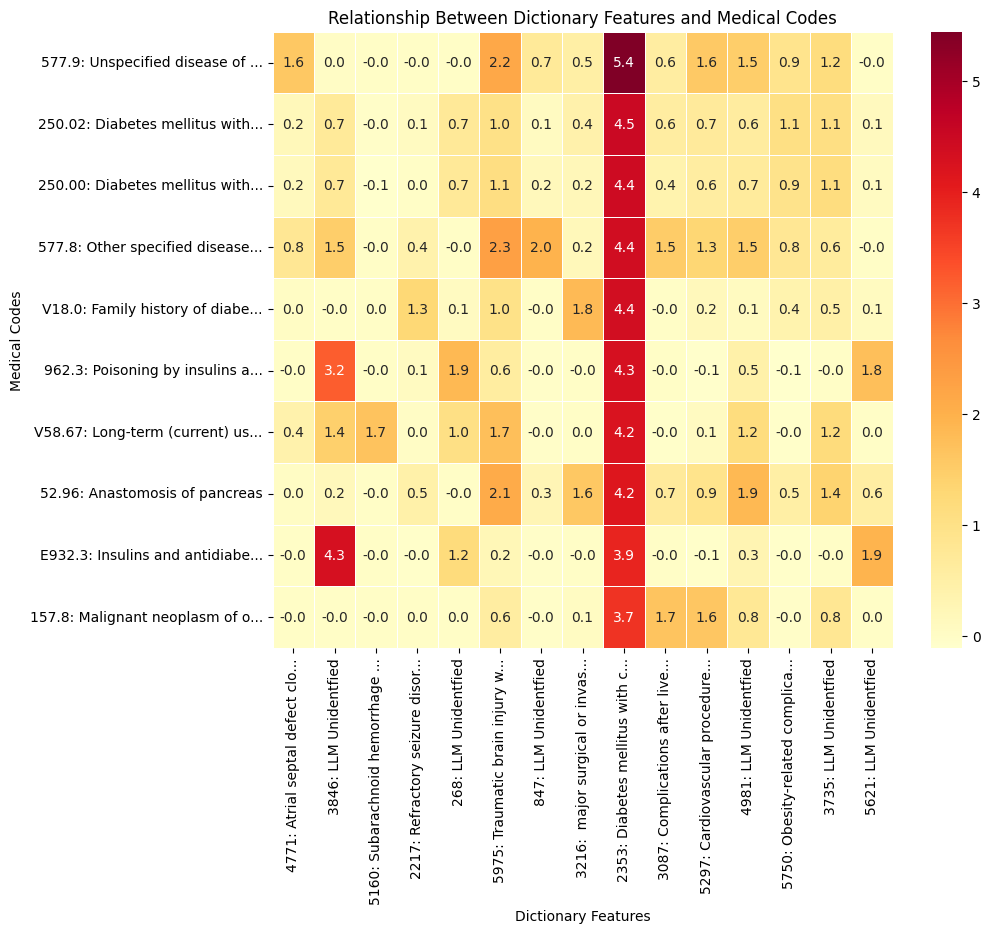}
\caption{Heat map visualizations of the $\mathbf{A_{f_\text{icd}}}$ matrix for medical codes associated with diabetes-related dictionary features. The x-axis represents the dictionary features, while the y-axis represents the respective medical codes. The intensity of each cell in the heat map indicates the strength of the association between a dictionary feature and a medical code. Dictionary features labeled as "LLM unidentified" denote instances where the LLM pipeline was unable to identify the underlying concept represented by the feature.}
\label{fig: f_heatmap_diabetes}
\end{figure}

From qualitative examinations of our human evaluations in Section \ref{Appendix : Human Predictability Experiment}, we notice that the model can often learn unintended associations between medical codes and specific medical conditions. We attempt to better visualize and discern other potential mismappings through heatmap visualizations. For instance, while they can be related, traumatic brain injuries aren't a direct cause of diabetic medical codes or conditions as observed in Figure \ref{fig: f_heatmap_diabetes}.

\begin{figure}[H] 
\centering
\includegraphics[width=1.0\textwidth]{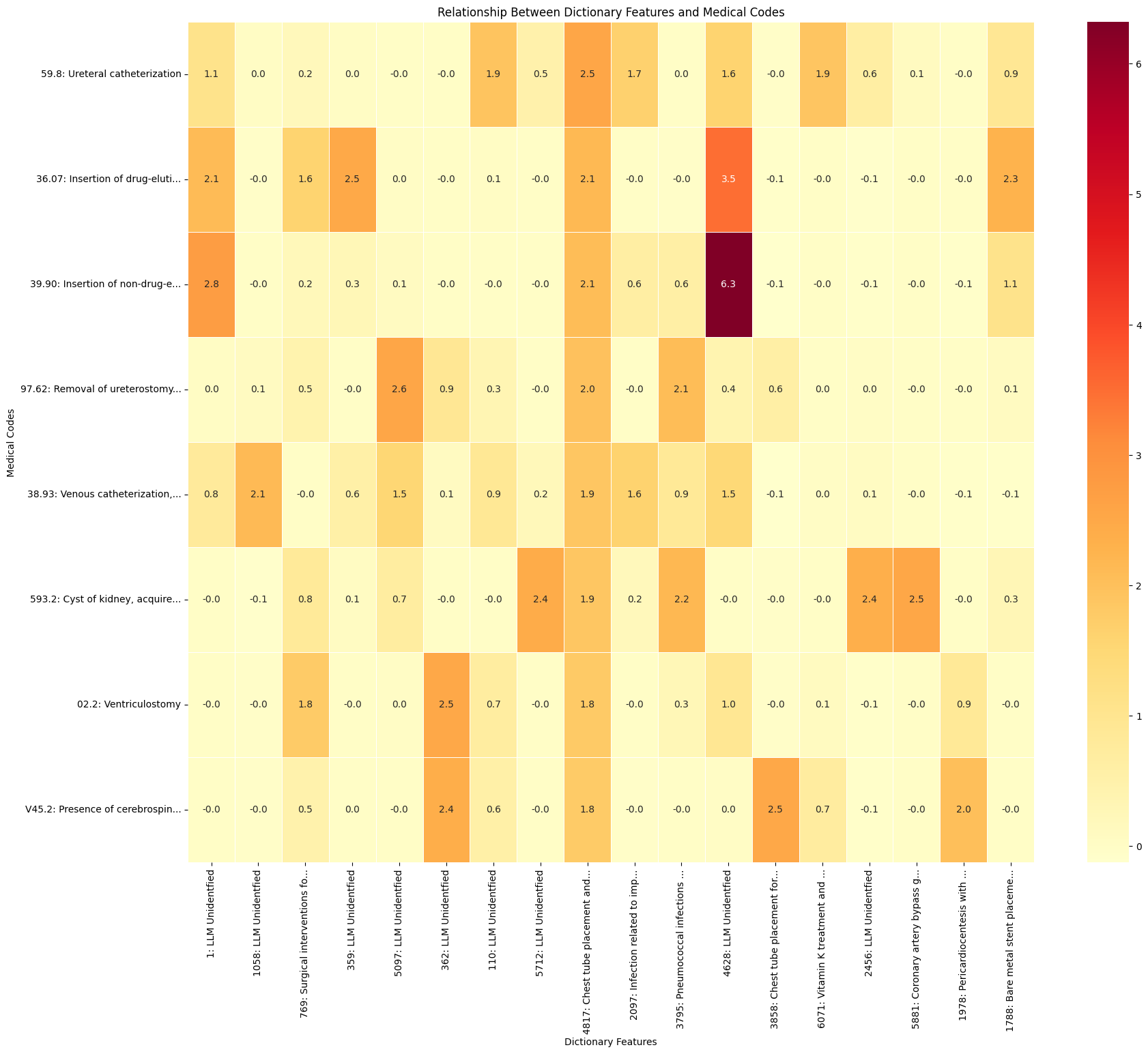}
\caption{Heat map visualizations of the $\mathbf{A_{f_\text{icd}}}$ matrix for medical codes associated with chest tube-related dictionary features. The x-axis represents the dictionary features, while the y-axis represents the respective medical codes. The intensity of each cell in the heat map indicates the strength of the association between a dictionary feature and a medical code. Dictionary features labeled as "LLM unidentified" denote instances where the LLM pipeline was unable to identify the underlying concept represented by the feature. }
\label{fig: f_heatmap_chest_tube}
\end{figure}

\begin{figure}[H] 
\centering
\includegraphics[width=1.0\textwidth]{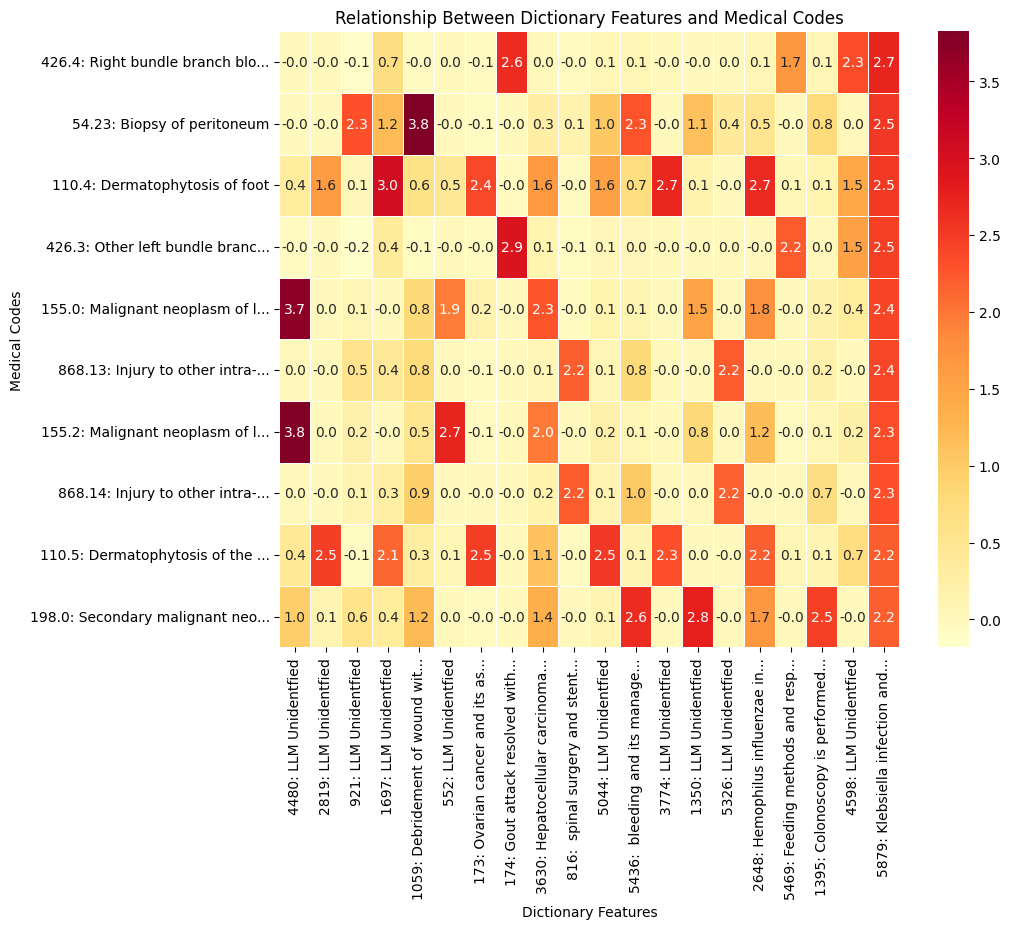}
\caption{Heat map visualizations of the $\mathbf{A_{f_\text{icd}}}$ matrix for medical codes associated with sepsis-related dictionary features. The x-axis represents the dictionary features, while the y-axis represents the respective medical codes. The intensity of each cell in the heat map indicates the strength of the association between a dictionary feature and a medical code. Dictionary features labeled as "LLM unidentified" denote instances where the LLM pipeline was unable to identify the underlying concept represented by the feature.}
\label{fig: f_heatmap_sepsis}
\end{figure}


\subsubsection{Debugging Case Study} \label{Appendix: Debugging}
\begin{figure}[H] 
\centering
\includegraphics[width=1.0\textwidth]{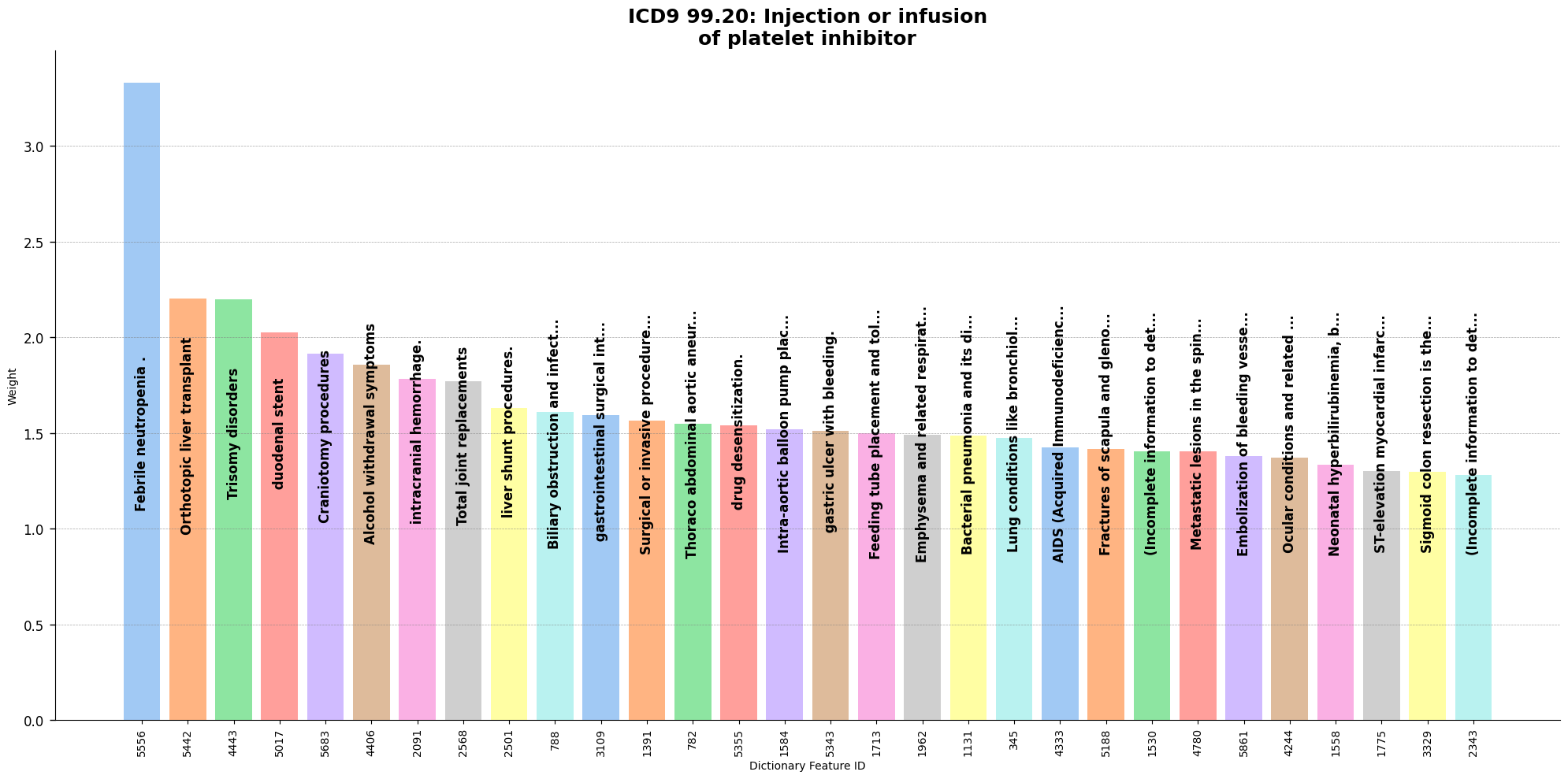}
\caption{Top 20 dictionary features associated with ICD 99.20 Injection or infusion of platelet inhibitors. We observe many unrelated dictionary features. Note that we also manually identify some of the previously LLM unidentified contexts. }
\label{fig: f_platelet_bar}
\end{figure}

Although there are more potentially, we showcase the top 20 different highly relevant dictionary features for the commonly falsely predicted ICD 99.20 code of "Injection of Platelet Inhibitors" in Figure \ref{fig: f_platelet_bar}. Digging deeper within the top 100 related dictionary features, we observe the following false dictionary features are not related to the medical code:
\begin{itemize}
\item 5188 Fractures of scapula and glenoid fossa observed repeatedly.
\item 4443 Trisomy disorders, likely Down syndrome (Trisomy 21), and associated genetic counseling.
\item 345 Lung conditions like bronchiolitis obliterans organizing pneumonia (BOOP) and radiation pneumonitis.
\item 1558 Neonatal hyperbilirubinemia, both physiologic and pathologic, and related diagnostic workup.
\item 802 Patients with end-stage amyotrophic lateral sclerosis (ALS) who are ventilator-dependent and have complications such as bronchiectasis, pneumonia, and cardiac issues.
\item 6069 Patients undergoing total knee replacement surgery, particularly those with complications like respiratory distress, pain management issues, or comorbidities such as schizophrenia.
\item 4917 Patients with severe dysphagia and difficulty managing oral secretions, often in the context of advanced illnesses such as cancer or critical conditions.
\end{itemize}

We ablate these dictionary features to showcase our initial debugging attempts.

\subsection{Unidentifiable Dictionary Features}
We attempt to investigate some of the highly relevant unidentifiable features in the heatmaps above and observe a couple findings. First, some dictionary features simply lack enough context, and second, some are truly challenging to discern a common pattern, often having a very diverse set of medical conditions that activate a specific dictionary feature.

\begin{table}[h]
\centering
\begin{tabular}{p{0.8\textwidth} p{0.2\textwidth}}
\toprule
Highly Activating Tokens & Dictionary Feature $f_i$ \\
\midrule
ograft & 3846 \\
\hline
Surgical procedure anterior pelvic ring external fixator posterior ring fixation, Pneum cephalus brain, External fixator posterior ring fixation sacro iliac screw, Temporal, Pneum cephalus, Posterior ring fixation sacro iliac screw suprapubic catheter placement present, Allergic anaphylaxis asthmaticus steroid, Steroids, Removal of external fixator leg, sarc 
flare & 362 \\
\hline
Smoker fasci & 1 \\
\hline
 main, e  coli, c, cardiac  catheterization, va ci, pulmonary, e  coli, defibrillator, catheterization& 3774\\
\hline
 i ab p, liver, i ab p, idiopathic, balloon & 552
\\
\hline
 axilla, v c,v c, line, fistulas, m v c, phal,infectious,port ath, perianal  fistulas & 1350 \\
\bottomrule
\end{tabular}
\caption{Examples of Unidentified Dictionary Features. Many lack the number of contexts, or some have very divergent highly activating contexts like e coli and cardiac catheterization.}
\label{tab: Unidentifiable Dictionary Features}
\end{table}
